\documentclass[10pt,twocolumn,letterpaper]{article}

\usepackage{titling}
\usepackage{cvpr}
\usepackage{times}
\usepackage{epsfig}
\usepackage{graphicx}
\usepackage{amsmath}
\usepackage{amssymb}

\usepackage{cite}
\usepackage{gensymb}
\usepackage{subcaption}
\usepackage{enumitem}
\usepackage[export]{adjustbox}
\usepackage{booktabs}
\usepackage[labelfont=bf]{caption}
\usepackage[hang,flushmargin]{footmisc} 
\usepackage{textpos}

\usepackage{tikz}
\usepackage{pgfplots}
\usepgfplotslibrary{groupplots}
\usetikzlibrary{matrix}
\pgfplotsset{equi/.style={
    only marks,
    color=orange,
    mark=triangle,
    line width=0.8mm,
  }
}

\pgfplotsset{cubemap/.style={
    only marks,
    color=black!20!green,
    mark=diamond,
    line width=0.8mm,
  }
}

\pgfplotsset{s2cnn/.style={
    only marks,
    color=red,
    mark=o,
    line width=0.8mm,
  }
}

\pgfplotsset{sphcnn/.style={
    only marks,
    color=blue,
    mark=square,
    line width=0.8mm,
  }
}

\pgfplotsset{spherenet/.style={
    color=black,
    mark=x,
    line width=0.8mm,
  }
}

\pgfplotsset{sphunet/.style={
    only marks,
    color=orange,
    mark=triangle*,
  }
}

\pgfplotsset{sphconv/.style={
    only marks,
    color=black!20!green,
    mark=diamond*,
  }
}

\pgfplotsset{projected/.style={
    color=red,
    mark=*,
  }
}

\pgfplotsset{ktn/.style={
    color=blue,
    mark=square*,
  }
}

\pgfplotsset{compat=1.9}

\usepackage[pagebackref=true,breaklinks=true,letterpaper=true,colorlinks,bookmarks=false]{hyperref}

\cvprfinalcopy 

\def\cvprPaperID{4887} 

\ifcvprfinal\fi
\begin{document}

\title{Kernel Transformer Networks for Compact Spherical Convolution}

\author{
Yu-Chuan Su\\
The University of Texas at Austin
\and
Kristen Grauman\\
Facebook AI Research\\
The University of Texas at Austin
}

\maketitle

\begin{abstract}
Ideally, $360\degree$ imagery could inherit the deep convolutional neural networks (CNNs) already trained with great success on perspective projection images.
However, existing methods to transfer CNNs from perspective to spherical images introduce significant computational costs and/or degradations in accuracy.
We present the Kernel Transformer Network (KTN) to efficiently transfer convolution kernels from perspective images to the equirectangular projection of $360\degree$ images.
Given a source CNN for perspective images as input,
the KTN produces a function parameterized by a polar angle and kernel as output.
Given a novel $360\degree$ image, that function in turn can compute convolutions for arbitrary layers and kernels as would the source CNN on the corresponding tangent plane projections.
Distinct from all existing methods, KTNs allow model transfer: the same model can be applied to different source CNNs with the same base architecture.
This enables application to multiple recognition tasks without re-training the KTN.
Validating our approach with multiple source CNNs and datasets, 
we show that KTNs improve the state of the art for spherical convolution.
KTNs successfully preserve the source CNN's accuracy, while offering transferability, scalability to typical image resolutions, and, in many cases, a substantially lower memory footprint\footnote{Code and data available at \url{http://vision.cs.utexas.edu/projects/ktn/}}.
\vspace*{-0.05in}
\end{abstract}

\begin{textblock*}{\textwidth}(0cm,-18cm)
\centering
In Proceedings of the IEEE Conference on Computer Vision and Pattern Recognition (CVPR), 2019.%
\end{textblock*}


\section{Introduction}

The $360\degree$ camera is an increasingly popular technology gadget, with sales expected to grow by $1500\%$ before 2022~\cite{360camera}.
As a result, the amount of $360\degree$ data is increasing rapidly.
For example, users uploaded more than a million $360\degree$ videos to Facebook in less than 3 years~\cite{fb360videostatistics}.
Besides videography, $360\degree$ cameras are also gaining attention for self-driving cars, automated drones, and VR/AR.
Because almost any application depends on semantic visual features,
this rising trend prompts an unprecedented need for visual recognition algorithms on $360\degree$ images.

\begin{figure}[t]
    \center
    \includegraphics[width=\linewidth]{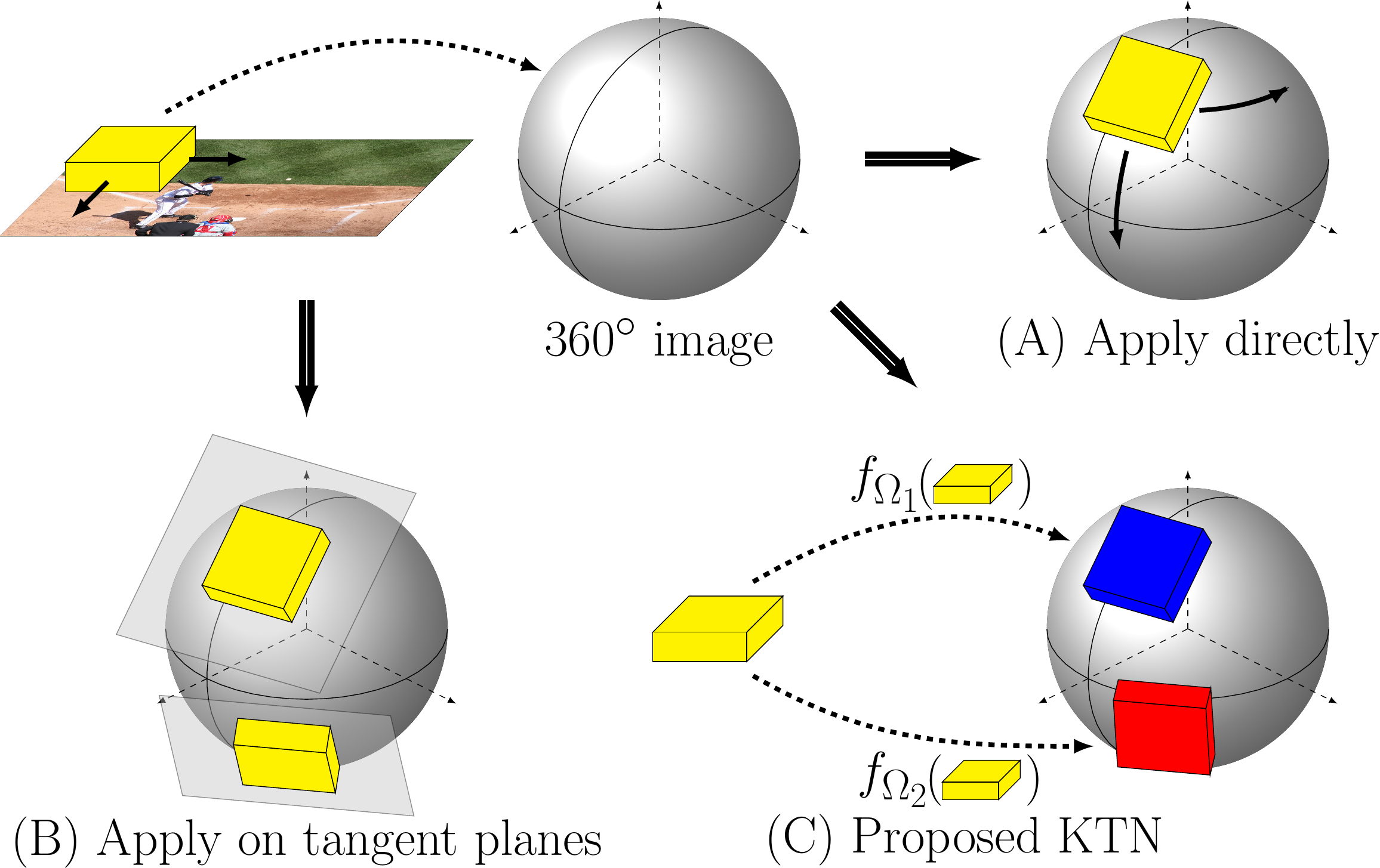}
    \\
    \caption[KTN]{
        Our goal is to transfer CNNs trained on planar images to $360\degree$ images.
        Common approaches either (A) apply CNNs directly on the equirectangular projection of a $360\degree$ image or (B) project the content to tangent planes and apply the models on the tangent planes.
        In contrast, Kernel Transformer Network (KTN) adapts the kernels in CNNs to account for the distortion in $360\degree$ images.
    }
    \vspace{-9pt}
    \label{fig:ktn}
\end{figure}

Today's wildly successful recognition CNNs are the result of tremendous data curation and annotation effort~\cite{imagenet,simonyan2014very,feichtenhofer2016convaction,tran2015learning,coco,deeplabv3plus2018},
but they all assume perspective projection imagery. 
How can they be repurposed for $360\degree$ data?
Existing methods often take an off-the-shelf model trained on perspective images and either
1) apply it repeatedly to multiple perspective projections of the $360\degree$ image~\cite{su2016accv,su2017cvpr,chou2017self,yu2018deepranking}
or 2) apply it once to a single equirectangular projection~\cite{lai2017semantic,hu2017deep}.
See Fig.~\ref{fig:ktn}(A,B).
These two strategies, however, have severe limitations.
The first is expensive because it has to project the image and apply the recognition model repeatedly.
The second is inaccurate because the visual content is distorted in equirectangular projection.

To overcome these challenges,
recent work designs CNN models specifically for spherical data~\cite{su2017nips,saliency360video,spherenet,cohen2017convolutional,sphericalcnn}.
Broadly speaking, they pursue one of three approaches.
The first adapts the network architecture for equirectangular projection and trains kernels of variable size to account for its distortions~\cite{su2017nips}.
While accurate, this approach suffers from significant model bloat.
The second approach instead adapts the kernels on the sphere, resampling the kernels or projecting their tangent plane features~\cite{saliency360video,spherenet}.
While allowing kernel sharing and hence smaller models, this approach degrades accuracy---especially for deeper networks---due to an implicit interpolation assumption, as we will explain below.
The third approach defines convolution in the spectral domain~\cite{cohen2017convolutional,sphericalcnn},
which has significant memory overhead and thus far limited applicability to real-world data.
All of the above require retraining to handle a new recognition task.

In light of these shortcomings,
we propose the Kernel Transformer Network (KTN).
The KTN adapts source CNNs trained on perspective images to $360\degree$ images.
Instead of learning a new CNN on $360\degree$ images for a specific task,
KTN learns a \emph{function} that takes a kernel in the source CNN as input and transforms it to be applicable to a $360\degree$ image in its equirectangular projection.
See Fig.~\ref{fig:ktn} (C).
The function accounts for the distortion in $360\degree$ images,
returning different transformations depending on both the polar angle $\theta$ and the source kernel.
The model is trained to reproduce the outputs of the source CNN on the perspective projection for each tangent plane on an arbitrary $360\degree$ image.
Hence, KTN learns to behave similarly to the source CNN while avoiding repeated projection of the image.

Key highlights of the proposed KTN are its \emph{transferability} and \emph{compactness}---both of which owe to our function-based design.
Once trained for a base architecture, the same KTN can transfer multiple source CNNs to $360\degree$ images.
For example, having trained a KTN for VGG~\cite{simonyan2014very} on ImageNet classification,
we can transfer the same KTN to run a VGG-based Pascal object detector on $360\degree$ panoramas.
This is possible because the KTN takes the source CNN as input rather than embed the CNN kernels into its own parameters (unlike~\cite{su2017nips,saliency360video,spherenet,cohen2017convolutional,sphericalcnn}). 
Furthermore, since the KTN factorizes source kernels from transformations, it is implementable with a lightweight network (e.g., increasing the footprint of a VGG network by only 25\%).

Results show KTN models are orders of magnitude smaller than the most accurate competitor, SphConv~\cite{su2017nips}.  
Compared with Spherical U-Net~\cite{saliency360video} and SphereNet~\cite{spherenet},
KTN is much more data efficient because it does not require any annotated $360\degree$ images for training,
and it is more accurate because it avoids their feature interpolation assumption.

\section{Related Work}

\paragraph{$360\degree$ vision}

Ongoing work explores new projection models optimized for
image display~\cite{zelnik2005squaring,chang-iccv2013,kim-iccv2017}
or video storage~\cite{fb2015cubemap,fb2016compressionrate,google2017eac,adeel2017rsp,su2018isomer}.
We adopt the most common equirectangular projection so our algorithm can be readily applied to existing data.
Other work explores how to improve the display of $360\degree$ video via video stabilization~\cite{kamali2011stabilizing,kasahara2015first,kopf2016tog}, new display interfaces~\cite{lin2017tell,pavel2017shot,lin2017outsidein},
and automatic view selection~\cite{su2016accv,lai2017semantic,hu2017deep,su2017cvpr,chou2017self,yu2018deepranking,cheng2018cubepadding}.
The latter all rely on applying CNNs to $360\degree$ data, and could benefit from our method.

\vspace{-8pt}
\paragraph{CNNs on spherical data}

As discussed above,
early methods take either the expensive but accurate reprojection approach~\cite{zhang2014panocontext,su2016accv,su2017cvpr,yu2018deepranking},
or the inaccurate but fast direct equirectangular approach~\cite{lai2017semantic,hu2017deep}.
Recent work improves accuracy by training vanilla CNNs on the cubemap projection, which introduces less distortion~\cite{boomsma2017nips,cheng2018cubepadding},
but the model still suffers from cubemap distortion and discontinuities and has sub-optimal accuracy for tasks such as object detection.

In the last year, several methods develop new spherical CNN models.
Some design CNN architectures that account for the distortion in $360\degree$ images~\cite{su2017nips,spherenet,saliency360video}.
SphConv~\cite{su2017nips} learns separate kernels for each row of the
equirectangular projection, training 
to reproduce the behavior of an off-the-shelf CNN and adjusting the kernel shape based on its location on the sphere.
While more accurate than a vanilla CNN,
SphConv increases the model size significantly because it unties kernel weights along the rows.
In contrast, SphereNet~\cite{spherenet} defines the kernels on the tangent plane and projects features to the tangent planes before applying the kernels.
Similarly, Spherical U-Net~\cite{saliency360video} defines the kernels on the sphere and resamples the kernels on the grid points for every location in the equirectangular projection.
Both allow weight sharing, 
but they implicitly assume that features defined on the sphere can be interpolated in the 2D plane defined by equirectangular projection, which we show is problematic.
Instead of learning independent kernels or using a fixed 2D transformation,
our KTN learns a transformation that considers both spatial and cross-channel correlation.
Our model is more compact than SphConv by sharing the kernels,
and it is more accurate than SphereNet and Spherical U-Net by learning a more generic transformation.

Another strategy is to define convolution in the spectral domain in order to learn rotation invariant CNNs.
One approach is to apply graph convolution and design the graph structure~\cite{khasanova2017graph} such that the outputs are rotation invariant.
Another approach transforms both the feature maps and kernels into the spectral domain and applies convolution there~\cite{cohen2017convolutional,sphericalcnn}.
However, orientation is often semantically significant in real data (e.g., cars are rarely upside down) and so removing orientation can unnecessarily restrict discrimination.
In addition, these approaches require caching the basis functions and the frequency domain feature maps in order to achieve efficient computation.
This leads to significant memory overhead and limits the viable input resolution.  
Both constraints limit the spectral methods' accuracy on real world $360\degree$ images.
Finally, unlike any of the above prior work~\cite{su2017nips,boomsma2017nips,cheng2018cubepadding,spherenet,saliency360video,cohen2017convolutional,sphericalcnn},
our KTN can transfer across different source CNNs with the same architecture to perform new tasks without re-training;
all other methods require training a new model for each task.

\vspace{-8pt}
\paragraph{CNNs with geometric transformations}

For perspective images, too, there is interest in encoding geometric transformations in CNN architectures.
Spatial transformer networks~\cite{jaderberg2015spatial} transform the feature map into a canonical view to achieve transformation invariance. 
Active convolution~\cite{jeon2017active} and deformable convolution~\cite{dai2017deformable} model geometric transformations using the receptive field of the kernel.
While these methods account for geometric transformations in the input data,
they are not suitable for $360\degree$ images because the transformation is \emph{location} dependent rather than \emph{content} dependent in $360\degree$ images.
Furthermore, all of them model only geometric transformation and ignore the correlation between different channels in the feature map.
In contrast, our method captures the properties of $360\degree$ images and the cross channel correlation in the features.

\section{Approach}
\label{sec:approach}

\setlength\abovedisplayskip{6pt}
\setlength\belowdisplayskip{6pt}

In this section, we introduce the Kernel Transformer Network for transferring convolutions to $360\degree$ images.
We first introduce the KTN module,
which can replace the ordinary convolution operation in vanilla CNNs.
We then describe the architecture and objective function of KTN.
Finally, we discuss the difference between KTN and existing methods for learning CNNs on $360\degree$ data.

\subsection{KTN for Spherical Convolution}

Our KTN can be considered as an generalization of ordinary convolutions in CNNs.
In the convolution layers of vanilla CNNs, the same kernel is applied to the entire input feature map to generate the output feature map.
The assumption underlying the convolution operation is that the feature patterns,
i.e., the kernels, are translation invariant and should remain the same over the entire feature map.
This assumption, however, does not hold in $360\degree$ images.
A $360\degree$ image is defined by the visual content projected on the sphere centered at the camera's optical center.
To represent the image in digital format, the sphere has to be unwrapped into a 2D pixel array, e.g., with equirectangular projection or cubemaps.  
Because all sphere-to-plane projections introduce distortion,
the feature patterns are not translation invariant in the pixel space,
and ordinary CNNs trained for perspective images do not perform well on $360\degree$ images.

To overcome this challenge,
we propose the Kernel Transformer Network, which can generate kernels that account for the distortion.
Assume an input feature map $I \in \mathbf{R}^{H \times W \times C}$ and a source kernel $K \in \mathbf{R}^{k \times k \times C}$ defined in undistorted images (i.e., perspective projection).
Instead of applying the source kernel directly
\begin{equation}
    \label{eq:ordinary_convolution}
    F[x, y] = \Sigma_{i,j} K[i, j] \ast I[x-i, y-j],
\end{equation}
we learn the KTN ($f$) that generates different kernels for different distortions:
\begin{align}
    K_{\Omega} &= f(K, \Omega) \label{eq:ktn}\\
    F[x, y] &= \Sigma_{i,j} K_{\Omega}[i, j] \ast I[x-i, y-j] \label{eq:ktn_convolution}
\end{align}
where the distortion is parameterized by $\Omega$.
Because the distortion in $360\degree$ images is location dependent,
we can define $\Omega$ as a function on the sphere
\begin{equation}
    \Omega = g(\theta, \phi),
\end{equation}
where $\theta$ and $\phi$ are the polar and azimuthal angle in spherical coordinates, respectively.
Given the KTNs and the new definition of convolution,
our approach permits applying an ordinary CNN to $360\degree$ images by replacing the convolution operation in Eq.~\ref{eq:ordinary_convolution} with Eq.~\ref{eq:ktn_convolution}.

KTNs make it possible to take a CNN trained for some target task (recognition, detection, segmentation, etc.) on ordinary perspective images and apply it directly to 360 panoramas.
Critically, KTNs do so without using any annotated $360\degree$ images.
Furthermore, as we will see below, once trained for a given architecture (e.g., VGG), the same KTN is applicable for a \emph{new} task using that architecture without retraining the KTN.
For example, we could train the KTN according to a VGG network trained for ImageNet classification,
then apply the same KTN to transfer a VGG network trained for Pascal object detection;
with the same KTN, both tasks can be translated to $360\degree$ images.

\subsection{KTN Architecture}

In this work, we consider $360\degree$ images that are unwrapped into 2D rectangular images using equirectangular projection.
Equirectangular projection is the most popular format for $360\degree$ images and is part of the $360\degree$ video compression standard~\cite{omaf2017wd}.
The main benefit of equirectangular projection for KTNs is that the distortion depends only on the polar angle.
Because the polar angle has an one-to-one correspondence with the image row ($y{=}\theta H / \pi$) in the equirectangular projection pixel space,
the distortion can be parameterized easily using $\Omega = g(\theta, \phi) = y$.
Furthermore, we can generate one kernel and apply it to the entire row instead of generating one kernel for each location,
which leads to more efficient computation.

A KTN instance is based on a given CNN architecture.  
There are two basic requirements for the KTN module.
First, it has to be lightweight in terms of both model size and computational cost.
A large KTN module would incur a significant overhead in both memory and computation,
which would limit the resolution of input $360\degree$ images during both training and test time.
Because $360\degree$ images by nature require a higher resolution representation in order to capture the same level of detail compared with ordinary images,
the accuracy of the model would degrade significantly if we were forced to use lower resolution inputs.

Second, KTNs need to generate output kernels with variable size,
because the appropriate kernel shape may vary in a single $360\degree$ image.
A common way to generalize convolution kernels on the 2D plane to $360\degree$ images is to define the kernels on the tangent plane of the sphere.
As a result, the receptive field of the kernel on the $360\degree$ image is the back projection of the receptive field on the tangent plane,
which varies at different polar angles~\cite{su2017nips,saliency360video,spherenet}.
While one could address this naively by always generating the kernels in the largest possible size,
doing so would incur significant overhead in both computation and memory.

We address the first requirement (size and cost) by employing depthwise separable convolutions~\cite{howard2017mobilenets,chollet2017xception} within the KTN.
Instead of learning 3D (i.e., height$\times$width$\times$channels) kernels,
KTN alternates between \emph{pointwise} convolution that captures cross-channel correlation and \emph{depthwise} convolution that captures spatial correlation.
Using the same 3x3 depthwise convolutions as in MobileNet~\cite{howard2017mobilenets},
the computation cost is about 8 to 9 times less than standard convolution.
Furthermore, the model size overhead for KTN is roughly $1/k^{2}$ of the source kernels,
where most of the parameters are in the 1x1 convolution.
The size overhead turns out to be necessary,
because cross channel correlation is captured only by the 1x1 convolution in KTN,
and removing it reduces the final spherical convolution accuracy significantly.

To address the second requirement (variable-sized kernels),
we learn a row dependent depthwise projection to resize the source kernel.
The projection consists of $h$ projection matrices $P_{i}$, for $i \in [1, h]$,
where $h$ is the number of rows in the $360\degree$ image.
Let $r_{i} = h_{i}\times w_{i}$ be the target kernel receptive field at row $i$.
The projection matrix has the size $P_{i} \in \mathbf{R}^{r_{i} \times k^{2}}$,
which projects the source kernel into the target size.
Similar to the depthwise convolution,
we perform channel-wise projection to reduce the model size.

\begin{figure}[t]
    \center
    \includegraphics[width=\linewidth]{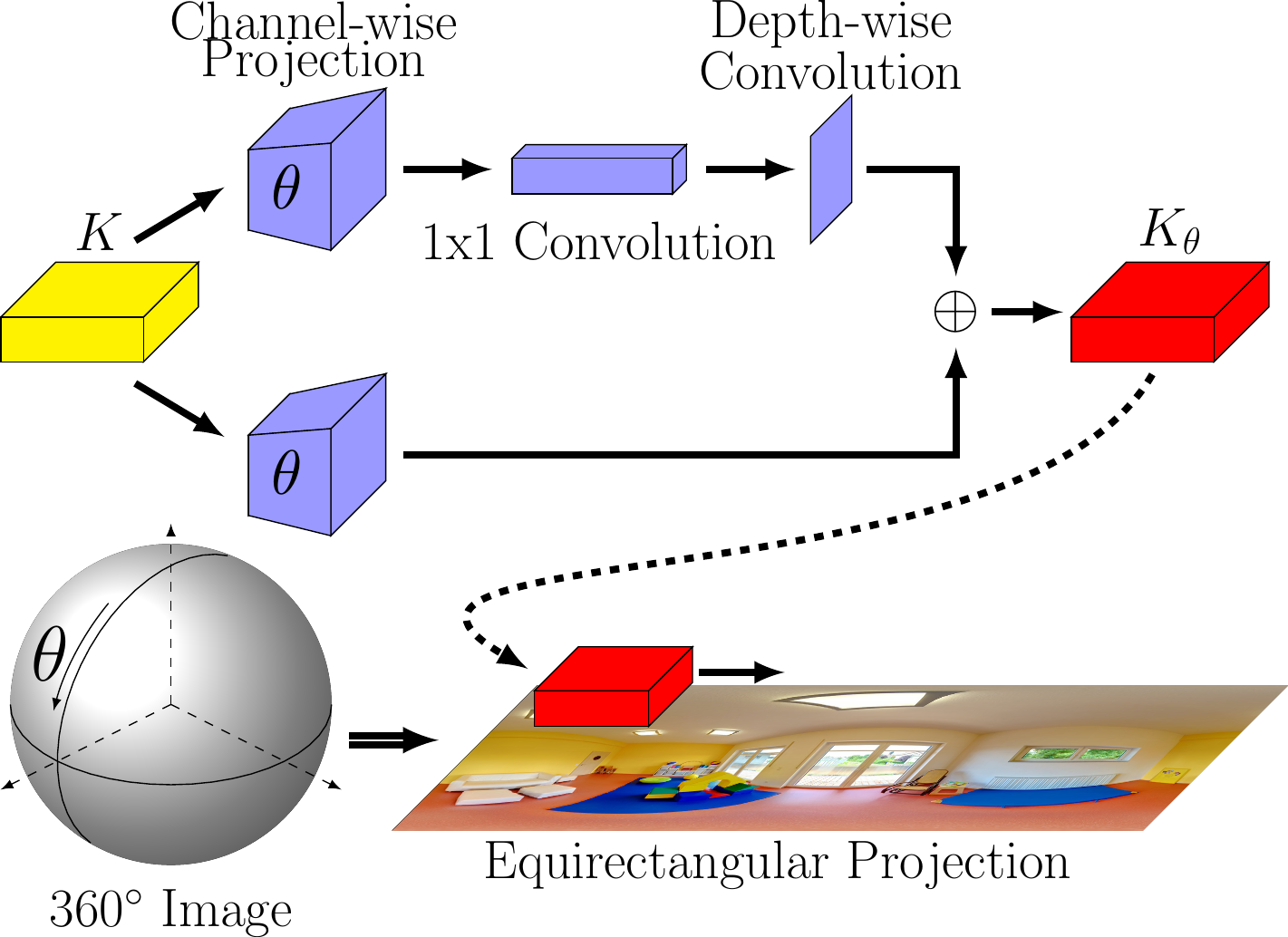}
    \\
    \vspace{-6pt}
    \caption[KTN architecture]{
        KTN consists of row dependent channel-wise projections that resize the kernel to the target size and depth separable convolution blocks.
        It takes a source kernel $K$ and $\theta$ as input and generates an output kernel $K_{\Omega}$.
        $K_{\Omega}$ is then applied to the $360\degree$ image in its equirectangular projection at row $y{=}\theta H/ \pi$.
        The transformation accounts for the distortion in equirectangular projection, while maintaining cross-channel interactions.
    }
    \label{fig:ktn_architecture}
    \vspace{-12pt}
\end{figure}

The complete architecture for KTN is in Fig.~\ref{fig:ktn_architecture}.
We use a Residual Network~\cite{he2016resnet}-like architecture.
For both the residual and shortcut branches,
we first apply the row dependent projection to resize the kernel to the target size.
The residual branch then applies depthwise separable convolution twice.
Our depthwise separable convolution block consists of ReLU-pointwise conv-ReLU-depthwise conv.
This design removes the batch normalization used in MobileNet to reduce the model size and memory consumption.
The two branches are added together to generate the output kernel,
which is then applied to a $360\degree$ feature map as in Eq.~\ref{eq:ktn_convolution}.
Note that while the KTN can be applied to different kernels,
the structure of a KTN depends on $P_{i}$, which is determined by the receptive field of the source kernel.
Therefore, we need one KTN for each layer of a source CNN.

\begin{table*}[t]\small
    \centering
    \caption{
        Comparison of different approaches.
        \textsc{Equirectangular} and \textsc{Cubemap} refer to applying the given CNN directly to the equirectangular and cubemap projection, respectively.
        Supervised training means that the method requires annotated 360 images.
        The model size is the size for a single layer,
        where $c, k, H$ refer to the number of channels, kernel size, and input resolution (bandwidth) respectively.
        Note that $c \sim H \gg k$ for real images and source CNNs, and we keep only the leading term for each method.
    } 
    \vspace{-12pt}
    \label{tab:approaches}
    \begin{tabular}{lcccccc}
                                                         & Translation & Rotation   & Supervised & Model            & Transferable \\
                                                         & Invariance  & Invariance & Training   & Size             & Across Models\\
        \midrule
        \textsc{Equirectangular}                         & No          & No         & No   & $c^{2}k^{2}$     & No  \\
        \textsc{Cubemap}                                 & No          & No         & No   & $c^{2}k^{2}$     & No  \\
        \textsc{$S^{2}$CNN}\cite{cohen2017convolutional} & Yes         & Yes        & Yes        & $c^{2}H $         & No  \\
        \textsc{Spherical CNN}\cite{sphericalcnn}        & Yes         & Yes        & Yes        & $c^{2}H $         & No  \\
        \textsc{Spherical U-Net}\cite{saliency360video}  & Yes         & No         & Yes        & $c^{2}k^{2}$     & No  \\
        \textsc{SphereNet}\cite{spherenet}               & Yes         & No         & Yes        & $c^{2}k^{2}$     & No  \\
        \textsc{SphConv}\cite{su2017nips}                & Yes         & No         & No         & $c^{2}k^{2}H $    & No  \\
        \midrule
        \textsc{KTN}                                     & Yes         & No         & No         & $c^{2}k^{2}+c^{2}$ & Yes \\
        \bottomrule
    \end{tabular}
    \vspace{-4pt}
\end{table*}

\subsection{KTN Objective and Training Process}

Having introduced the KTN module and how to apply it for CNNs on $360\degree$ images, we now describe the KTN objective function and training process.
The goal of the KTN is to adapt the source kernel to the $360\degree$ domain.  
Therefore, we train the model to reproduce the outputs of the source kernels.
Let $F^{l} \in \mathbf{R}^{H \times W \times C^{l}}$ and $F^{l+1} \in \mathbf{R}^{H \times W \times C^{l+1}}$ be the feature maps generated by the $l$-th and $(l+1)$-th layer of a source CNN respectively.
Our goal is to minimize the difference between the feature map generated by the source kernels $K^{l}$ and that generated by the KTN module:
\begin{equation}
    \mathcal{L} = \|F^{l+1} - f^{l}(K^{l}, \Omega) \ast F^{l}\|^{2}
    \label{eq:objective}
\end{equation}
for any $360\degree$ image.
Note that during training the feature maps $F^{l}$ are not generated by applying the source CNN directly on the equirectangular projection of the $360\degree$ images.
Instead, for each point $(x, y)$ in the $360\degree$ image,
we project the image content to the tangent plane of the sphere at
\begin{equation}
    (\theta, \phi) = (\frac{\pi \times y}{H}, \frac{2\pi \times x}{W})
\end{equation}
and apply the source CNN on the tangent plane.
This ensures that the target training values are accurately computed on undistorted image content.
$F^{l}[x, y]$ is defined as the $l$-th layer outputs generated by the source CNN at the point of tangency.
Our objective function is similar to that of SphConv~\cite{su2017nips},
but, importantly, we optimize the model over the entire feature map instead of on a single polar angle in order to factor the kernel itself out of the KTN weights.

The objective function depends only on the source pretrained CNN and does not require any annotated data for training.
In fact, it does not require image data specific to the target task,
because the loss is defined over $360\degree$ images.
In practice, we sample arbitrary $360\degree$ images for training regardless of the source CNN. 
For example, in experiments we train a KTN on YouTube video frames and then apply it for a Pascal object detection task.
Our goal is to fully reproduce the behavior of the source kernel.
Therefore, even if the training images do not contain the same objects, scenes, etc.~as are seen in the target task,
the KTN should still minimize the loss in Eq.~\ref{eq:objective}.
Although KTN takes only the source kernels and $\theta$ as input,
the exact transformation $f$ may depend on all the feature maps $F^{l},F^{l-1},\ldots,F^{1}$ to resolve the error introduced by non-linearities.
Our KTN learns the important components 
of those transformations from data. 
KTN's transferability across source kernels is analogous to the generalizability of visual features across natural images.
In general, the more visual diversity in the unlabeled training data, the more accurately we can expect the KTN to be trained.
While one could replace all convolution layers in a CNN with KTNs and train the entire model end-to-end using annotated $360\degree$ data,
we believe that Eq.~\ref{eq:objective} is a stronger condition while also enjoying the advantage of bypassing any annotated training data.

\subsection{Discussion}
\label{sub:discussion}

Compared to existing methods for convolution for $360\degree$ images,
the main benefits of KTN are its \emph{compactness} and \emph{transferability}.
The information required to solve the target task is encoded in the source kernel,
which is fed into the KTN as an \emph{input} rather than part of the model.
As a result, the same KTN can be applied to another CNN having the same base architecture but trained for a different target task.
In other words, without additional training,
the same KTN model can be used to solve multiple vision tasks on $360\degree$ images by replacing the source kernels,
provided that the source CNNs for each task have the same base architecture.

Most related to our work is the spherical convolution approach (SphConv)~\cite{su2017nips}.
SphConv learns the kernels adapted to the distortion in equirectangular projection.
Instead of learning the transformation function $f$ in Eq.~\ref{eq:ktn},
SphConv learns $K_{\Omega}$ directly, and hence must learn one $K_{\Omega}$ for every different row of the equirectangular image.
While SphConv should be more accurate than KTN theoretically (i.e., removing any limitations on memory and training time and data)
our experimental results show that the two methods perform similarly in terms of accuracy.
Furthermore, the number of parameters in SphConv is hundreds of times larger than KTN,
which makes  SphConv much more difficult to train and deploy.
The difference in model size becomes even more significant when there are multiple models to be evaluated:
the same \textsc{KTN} can apply to multiple source CNNs and thus incurs only constant overhead,
whereas SphConv must fully retrain and store a new model for each source CNN.
For example, if we want to apply five different VGG-based CNNs to $360\degree$ images,
SphConv will take $29{\times}5{=}145$GB of space,
while \textsc{KTN} takes only $56{\times}5{+}14{=}294$MB (cf.~Sec.~\ref{sec:cost}).
In addition, since SphConv trains $K_{\Omega}$ for a single source kernel $K$,
the model does not generalize to different source CNNs.

\begin{figure}[t]
    \center
    \includegraphics[width=.8\linewidth]{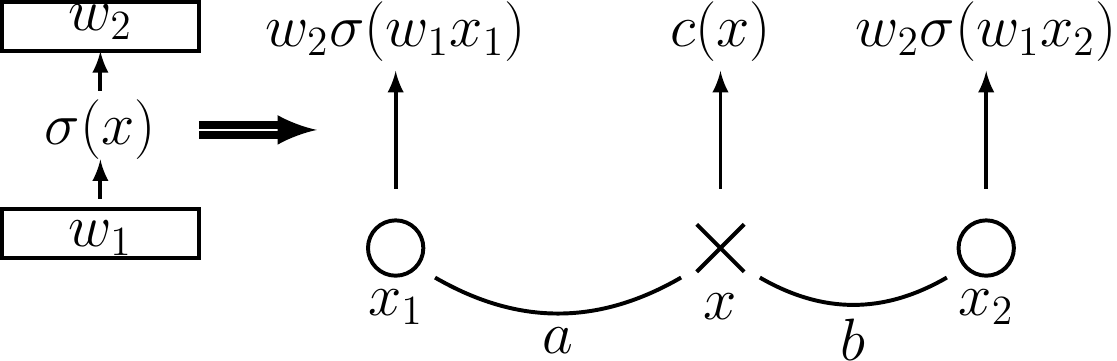}
    \\
    \vspace{-4pt}
    \caption[Convolution]{
        Beyond the first CNN layer, the feature interpolation assumption in SphereNet~\cite{spherenet} yields only approximated results.  See text for details.
    }
    \label{fig:convolution_example}
    \vspace{-12pt}
\end{figure}

SphereNet~\cite{spherenet} formulates the transformation function $f$ using the sphere-to-tangent-plane image projection.
While the projection transformation leads to an analytical solution for $f$, 
it implicitly assumes that CNN feature maps can be interpolated like pixels.
This assumption is only true for the first layer in a network because of non-linear activation functions used in modern CNNs between convolution layers.
Consider a two layer 1D convolution with a kernel of size 1, as sketched in Fig.~\ref{fig:convolution_example}.
If we interpolate the pixel first and apply the kernels,
the output of at location $x$ is
\begin{equation}
   c(x) =  w_{2} \times \sigma(w_{1} (a x_{1} + b x_{2})).
\end{equation}
However, if we apply the kernels and then interpolate the features,
the result is
\begin{equation}
   c(x) =  a w_{2} \times \sigma(w_{1} x_{1}) + b w_{2} \times \sigma(w_{1} x_{2}).
\end{equation}
These two values are not equal because $\sigma$ is non-linear,
and the error will propagate as the network becomes deeper.
The interpolated feature can at most be an approximation for the exact feature.
Our experimental results show that a projection transformation for $f$ leads to sub-optimal performance.

Finally, other methods attempt to reduce distortion by unwrapping a single $360\degree$ image into multiple images using perspective projection locally~\cite{boomsma2017nips,cheng2018cubepadding}, e.g., with cubemap projection.
It is non-trivial to define convolution across multiple image planes, where two cube faces meet.
Prior work addresses this problem by ``cube-padding'' the feature maps using output from adjacent image planes~\cite{boomsma2017nips,cheng2018cubepadding},
but experimental results indicate that the resultant features are not accurate enough and degrade the accuracy.
The reason is that the same object may have different appearance on different tangent planes,
especially when the field-of-view is large and introduces significant perspective distortion.
Alternatively, one could sample the tangent planes densely and apply convolution on each tangent plane independently, but
doing so incurs unrealistic computational overhead~\cite{su2017cvpr}.

Table~\ref{tab:approaches} summarizes the tradeoffs between existing spherical convolution models.
In short, KTN is distinct from all others in its ability to transfer to new tasks without any labeled data.
Furthermore, KTN has the favorable properties of a highly compact model and the ability to preserve orientation-specific features (typically desirable for recognition and other high-level tasks).

\section{Experiments}
\label{sec:experiments}

We evaluate KTN on multiple datasets and multiple source models.
The goal is to
1) validate the accuracy of KTN as compared to other methods for learning CNNs on $360\degree$ images,
2) demonstrate KTN's ability to generalize to novel  source models,
and 3) examine KTN's memory and computation overhead compared to existing techniques.

\vspace{-8pt}
\paragraph{Datasets}
\label{par:datasets}

Our experiments make use of both unannotated $360\degree$ videos and $360\degree$ images with annotation. 

\emph{Spherical MNIST} is constructed from the MNIST dataset by back projecting the digits into equirectangular projection with $160{\times}80$ resolution.
The digit labels are used to train the source CNN (recognition model), but they are \emph{not} used to train the KTN.
Classification accuracy on the $360\degree$-ified test set is used as the evaluation metric. 

\emph{Pano2Vid} is a real world $360\degree$ video dataset~\cite{su2016accv}.
We sample frames from non-overlapping videos for training and testing, and the frames are resized to $640{\times}320$ resolution.
The models are trained to reproduce the convolution outputs of the source model, so no labels are required for training.
The root-mean-square error (RMSE) of the final convolution outputs is used as the evaluation metric.

\emph{Pascal VOC 2007} is a perspective image dataset with object annotations.
We backproject the object bounding boxes to equirectangular projection with $640{\times}320$ resolution.
Following~\cite{su2017nips}, we use the accuracy of the detector network in Faster R-CNN on the validation set as the evaluation metric.
This dataset is used for evaluation only.

\vspace{-8pt}
\paragraph{Source Models}
\label{par:models}

For \emph{Spherical MNIST}, we train the source CNN on the MNIST training set.
The model consists of three convolution layers followed by one fully connected layer.
Each convolution layer consists of 5x5Conv-MaxPool-ReLU,
and the number of kernels is 32, 64, and 128, respectively.
For \emph{Pano2Vid} and \emph{Pascal VOC},
we take off-the-shelf Faster R-CNN~\cite{ren2015fasterRCNN} models with VGG architecture~\cite{simonyan2014very} as the source model.
The Faster R-CNN is trained on Pascal VOC if not mentioned specifically.
Source models are not fine-tuned on $360\degree$ data in any form.

\vspace{-8pt}
\paragraph{Baselines}
\label{par:baselines}
We compare to the following existing methods:
\begin{itemize}[leftmargin=*,label=$\bullet$]
    \setlength{\itemsep}{1pt}
    \setlength{\parskip}{1pt}
    \item \textsc{Equirectangular}---Apply ordinary CNNs on the $360\degree$ image in its equirectangular projection.
    \item \textsc{Cubemap}---Apply ordinary CNNs on the $360\degree$ image in its cubemap projection.
    \item \textsc{$S^{2}$CNN}~\cite{cohen2017convolutional}---We train \textsc{$S^{2}$CNN} using the authors' implementation.
        For \emph{Pano2Vid} and \emph{Pascal VOC}, we reduce the input resolution to $64{\times}64$ due to memory limits (see Supp).
        We add a linear read-out layer at the end of the model to generate the final feature map.
    \item \textsc{Spherical CNN}~\cite{sphericalcnn}---We train \textsc{Spherical CNN} using the authors' implementation.
        Again, the resolution of input is scaled down to $80{\times}80$ due to memory limits for \emph{Pano2Vid} and \emph{Pascal VOC}.
    \item \textsc{Spherical U-Net}~\cite{saliency360video}---We use the spherical convolution layer in Spherical U-Net to replace ordinary convolution in CNN.
        Input resolution is reduced to $160{\times}80$ for \emph{Pano2Vid} and \emph{Pascal VOC} due to memory limits.
    \item \textsc{SphereNet}~\cite{spherenet}---We implement \textsc{SphereNet} using row dependent channel-wise projection.\footnote{The authors' code and data were not available at the time of publication.}
        We derive the weights of the projection matrices using the feature projection operation and train the source kernels.
        For the \emph{Pano2Vid} dataset, 
        we train each layer independently using the same objective as \textsc{KTN} due to memory limits.
    \item \textsc{SphConv}~\cite{su2017nips}---We use the authors' implementation.
    \item \textsc{Projected}---Similar to \textsc{SphereNet}, except that it uses the source kernels without training.
\end{itemize}

The network architecture for \textsc{Equirectangular} and \textsc{Cubemap} is the same as the source model.
For all methods, the number of layers and kernels are the same as the source model.

Note that the resolution reductions specified above were necessary to even run those baseline models on the non-MNIST datasets, even with state-of-the-art GPUs.
All experiments were run on NVIDIA V100 GPU with 16GB memory---the largest in any generally available GPU today.
Therefore, the restriction is truly imposed by the latest hardware technology.
Compatible with these limits, the resolution in the authors' own reported results is restricted to $60\times60$~\cite{cohen2017convolutional}, $64\times64$~\cite{sphericalcnn}, or $150\times300$~\cite{saliency360video}.
On the \emph{SphericalMNIST} dataset, all methods use the exact same image resolution.
The fact that KTN scales to higher resolutions is precisely one of its technical advantages, which we demonstrate on the other datasets. 

For \emph{Spherical MNIST}, the baselines are trained to predict the digit projected to the sphere except \textsc{SphConv}.
\textsc{SphConv} and our \textsc{KTN} are trained to reproduce the conv3 outputs of the source model.
For \emph{Pano2Vid}, all methods are trained to reproduce the conv5\_3 outputs.

Please see Supp.~file for additional details.

\subsection{Model Accuracy}

\begin{table}[t]\small
    \vspace{-4pt}
    \centering
    \caption{Model accuracy.}
    \vspace{-8pt}
    \label{tab:accuracy}
    \begin{tabular}{lccc}
      & \emph{MNIST} & \emph{Pano2Vid} & \emph{Pascal VOC} \\
      &     (Acc.$\uparrow$) & (RMSE $\downarrow$) &  (Acc.$\uparrow$)\\
        \midrule
        \textsc{Equirectangular}                         & 95.24          & 3.44            & 41.63 \\
        \textsc{Cubemap}                                 & 68.53          & 3.57            & 49.29 \\
        \midrule
        \textsc{$S^{2}$CNN}\cite{cohen2017convolutional} & 95.79          & 2.37                  & 4.32 \\
        \textsc{Spherical CNN}\cite{sphericalcnn}        & 97.48          & 2.36            & 6.06 \\
        \textsc{Spherical U-Net}\cite{saliency360video}  & 98.43          & 2.54                  & 24.98 \\
        \textsc{SphereNet}\cite{spherenet}               & 87.20          & 2.46                  & 46.68 \\
        \textsc{SphConv}\cite{su2017nips}                & \textbf{98.72} & \textbf{1.50}   & 63.54 \\
        \textsc{Projected}                               & 10.70          & 4.24            & 6.15\\
        \midrule
        \textsc{KTN}                                     & 97.94          & 1.53            & \textbf{69.48} \\
        \bottomrule
    \end{tabular}
    \vspace{-4pt}
\end{table}

Table~\ref{tab:accuracy} summarizes the methods' CNN accuracy on all three $360\degree$ datasets.
\textsc{KTN} performs on par with the best baseline method (\textsc{SphConv}) on \emph{Spherical MNIST}.
The result verifies that \textsc{KTN} can transfer the source kernels to the entire sphere by learning to reproduce the feature maps,
and it can match the accuracy of existing models trained with annotated $360\degree$ images.

\textsc{KTN} and \textsc{SphConv} perform significantly better than the other baselines on the high resolution datasets,
i.e., \emph{Pano2Vid} and \emph{Pascal VOC}.
\textsc{$S^{2}CNN$}, \textsc{Spherical CNN}, and \textsc{Spherical U-Net} suffer from their memory constraints, which as discussed above restricts them to lower resolution inputs.
Their accuracy is significantly worse on realistic full resolution datasets.
These models cannot take higher resolution inputs even after using model parallelism over four GPUs with a total of 64GB of memory.
Although \textsc{Equirectangular} and \textsc{Cubemap} are trained and applied on the full resolution inputs,
they do not account for the distortion in $360\degree$ images and yield lower accuracy.
Finally, the performance of \textsc{Projected} and \textsc{SphereNet} suggests that the transformation $f$ cannot be modeled by a tangent plane-to-sphere projection.
Although \textsc{SphereNet} shows that the performance can be significantly improved by training the source kernels on $360\degree$ images,
the accuracy is still worse than \textsc{KTN} because feature interpolation introduces error.
The error accumulates across layers, as discussed in Sec.~\ref{sub:discussion},
which substantially degrades the accuracy when applying a deep CNN.
Note that the number of learnable parameters in \textsc{KTN} is much smaller than that in \textsc{SphereNet},
but it still achieves a much higher accuracy.

\begin{figure}[t]
    \vspace{-4pt}
    \centering
    \includegraphics[width=.49\linewidth]{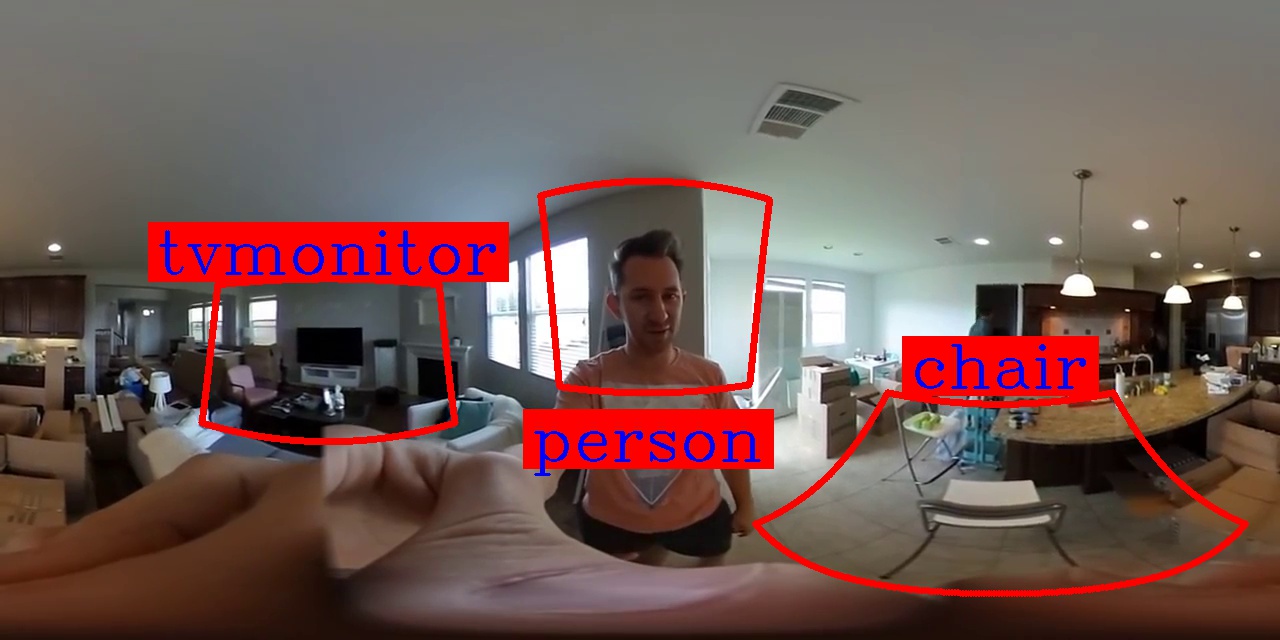}
    \includegraphics[width=.49\linewidth]{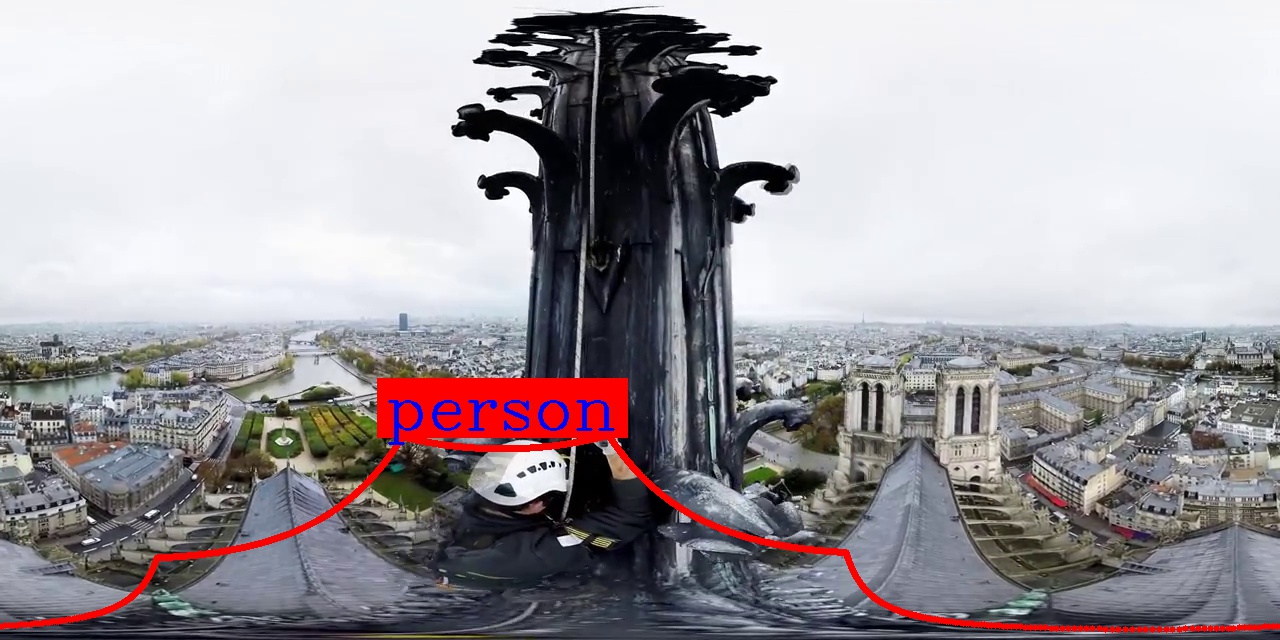}\\
    \includegraphics[width=.49\linewidth]{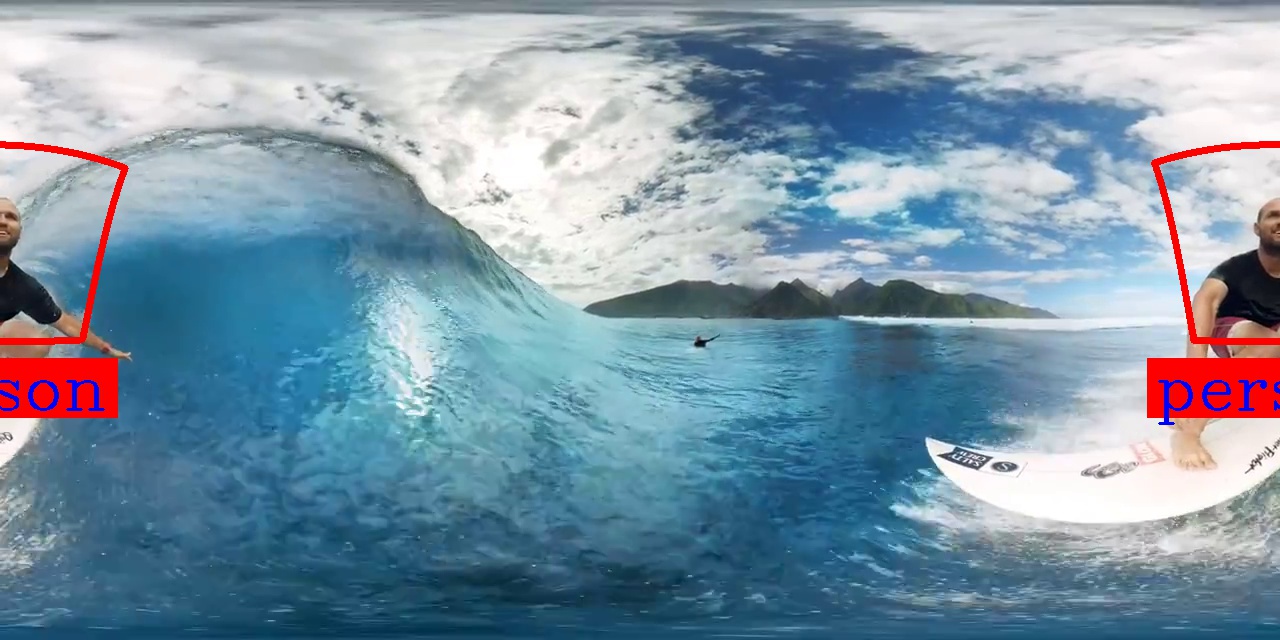}
    \includegraphics[width=.49\linewidth]{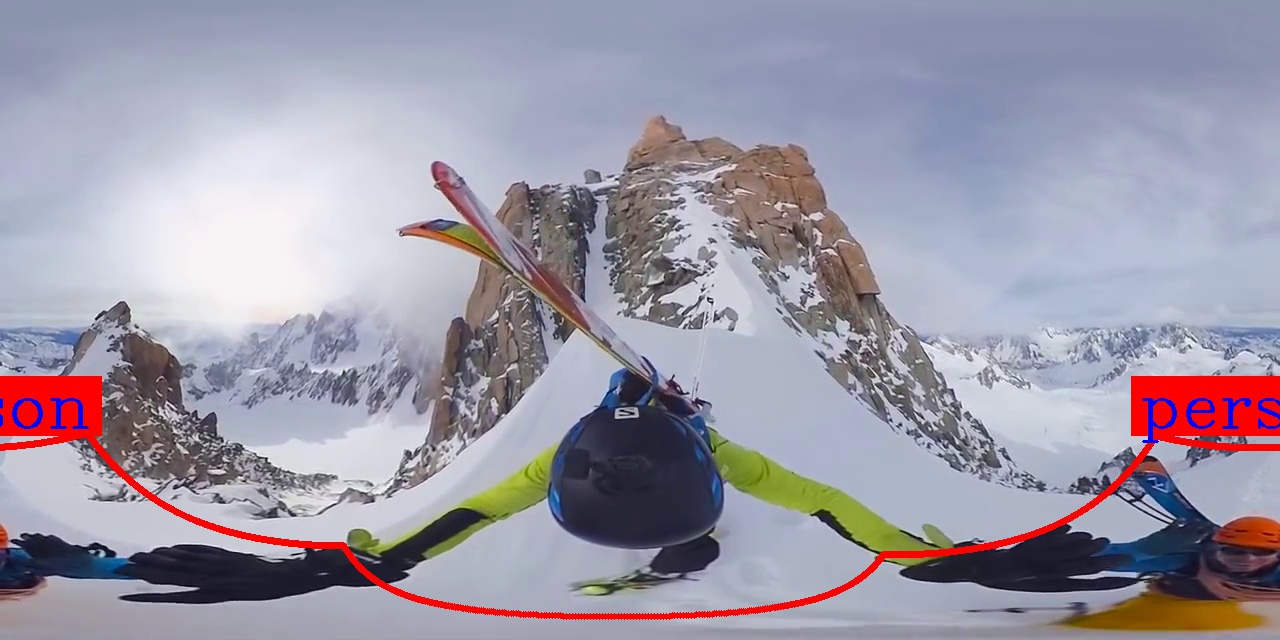}\\
    \vspace{-4pt}
    \caption{
        \textsc{KTN} object detection examples on \emph{Pano2Vid}.
        See Supp.~for detection examples on \emph{Pascal VOC}.
    }
    \label{fig:qualitative}
    \vspace{-4pt}
\end{figure}

Interestingly,
although \textsc{SphConv} performs better in RMSE on \emph{Pano2Vid},
\textsc{KTN} peforms better in terms of object classification accuracy on \emph{Pascal VOC}.
We attribute this to \textsc{KTN}'s inherent generalizability.
\textsc{SphConv} has a larger number of parameters,
and the kernels at different $\theta$ are trained independently.
In contrast, the parameters in \textsc{KTN} are shared across different $\theta$ and thus trained with richer information.
Therefore, \textsc{SphConv} is more prone to overfit the training loss,
which is to minimize the RMSE for both models.
Furthermore, our \textsc{KTN} has a significant compactness advantage over \textsc{SphConv}, as discussed above.

Similarly,
although \textsc{Spherical U-Net} and \textsc{SphereNet} perform slightly worse than \textsc{$S^{2}CNN$} and \textsc{Spherical CNN} on \emph{Pano2Vid},
they are significantly better than those baselines on \emph{Pascal VOC}.
This result reinforces the practical limitations of imposing rotation invariance.
\textsc{$S^{2}CNN$} and \textsc{Spherical CNN} require full rotation invariance; 
the results show that orientation information is in fact important in tasks like object recognition.
Thus, the additional rotational invariance constraince limits the expressiveness of the kernels and degrades the performance of \textsc{$S^{2}CNN$} and \textsc{Spherical CNN}.
Furthermore, the kernels in \textsc{$S^{2}CNN$} and \textsc{Spherical CNN} may span the entire sphere,
whereas spatial locality in kernels has proven important in CNNs for visual recognition.

Fig.~\ref{fig:qualitative} shows example outputs of \textsc{KTN} with a Faster R-CNN source model.
The detector successfully detects objects despite the distortion.
On the other hand, KTN can fail when a very close object cannot be captured in the field-of-view of perspective images.

\subsection{Transferability}

\begin{figure}[t]
    \vspace{-4pt}
    \center
    \includegraphics[width=\linewidth]{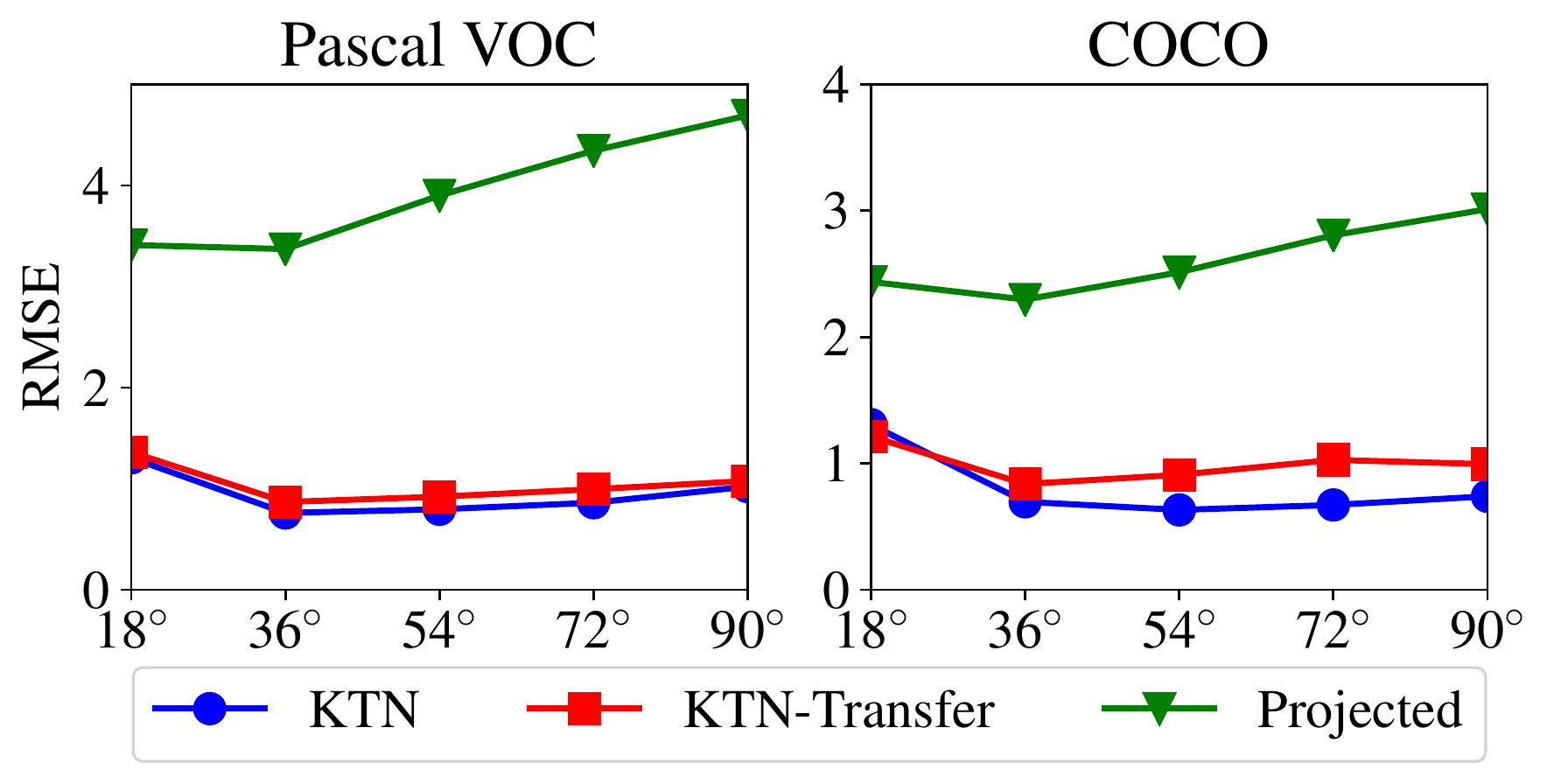}
    \vspace{-18pt}
    \caption{
        Model transferability.
        The title indicates the source CNN being tested.
        \textsc{KTN} performs almost identically regardless of the source network it is trained on.
        The results show we can learn a single \textsc{KTN} and apply it to other source CNNs with the same architecture,
        even if that source model is trained for a different task.
    }
    \label{fig:transferability}
    \vspace{-8pt}
\end{figure}

Next, we evaluate the transferability of \textsc{KTN} across different source models on \emph{Pano2Vid}.
In particular,
we evaluate whether KTNs trained with a Faster R-CNN that is trained on COCO can be applied to another Faster R-CNN (both using VGG architecture) that is trained on Pascal VOC and vice versa.
We denote KTN trained on a different source CNN than it is being tested on as \textsc{KTN-transfer} and \textsc{KTN} otherwise.

Fig.~\ref{fig:transferability} shows the results.
The accuracy of \textsc{KTN-Transfer} is almost identical to \textsc{KTN}.
The results demonstrate that \textsc{KTN} indeed learns a task-independent transformation and can be applied to different source models with the same base architecture.
None of the existing models~\cite{cohen2017convolutional,sphericalcnn,saliency360video,su2017nips,spherenet} are equipped to perform this kind of transfer,
because they learn fixed kernels for a specific task in some form.
Hence, the \textsc{Projected} baseline is the only baseline shown in Fig.~\ref{fig:transferability}.
Although \textsc{Projected} can be applied to any source CNN without training, the performance is significantly worse than \textsc{KTN}.
Again, the results indicate that a projection operation is not sufficient to model the required transformation $f$.
The proposed \textsc{KTN} is the first approach to spherical convolution that translates across models without requiring labeled $360\degree$ images or retraining.
We also perform the same experiments between VGG trained for ImageNet classification and Faster R-CNN trained for Pascal object detection,
and the results are similar. See Supp.

\subsection{Size and Speed}

\label{sec:cost}

Finally, we compare the overhead introduced by \textsc{KTN} versus that required by the baseline methods.
In particular, we measure the model size and speed for the convolution layers in the VGG architecture.
For the model size,
we compute the total size of the parameters using 32-bit floating point numbers for the weights.
While there exist algorithms that compress neural networks, they are equally applicable for all methods.
For the speed, we measure the average processing time (I/O excluded) of an image for computing the conv5\_3 outputs.
All methods are evaluated on a dedicated AWS p3.8xlarge instance.
Because the model size for \textsc{SphConv} is 29GB and cannot fit in GPU memory (16GB), it is run on CPUs.
Other methods are run on GPUs.

\begin{figure}[t]
    \vspace{-4pt}
    \center
\resizebox{\linewidth}{!} {
    \begin{tikzpicture}
        \tikzset{mark options={mark size=4, solid}}
        \begin{groupplot}[
            group style={group size=1 by 2,
                         horizontal sep=3em,
                         vertical sep=3em},
            width=0.25\textwidth,
            axis lines=left,
        ]
            \nextgroupplot[
                ymin=0,
                ymax=80,
                xmin=50,
                xmax=10000,
                xmode=log,
                only marks,
                ylabel={\large Accuracy},
                xlabel={\large Size (MB)},
            ]
                \addplot[equi] coordinates {(56.13, 41.63)};
                \label{plots:equi}
                \addplot[cubemap] coordinates {(56.13, 49.29)};
                \label{plots:cubemap}
                \addplot[s2cnn] coordinates {(464.14, 4.32)};
                \label{plots:s2cnn}
                \addplot[sphcnn] coordinates {(129.41, 6.06)};
                \label{plots:sphcnn}
                \addplot[spherenet] coordinates {(56.13, 46.68)};
                \label{plots:spherenet}
                \addplot[sphunet] coordinates {(56.15, 24.98)};
                \label{plots:sphunet}
                \addplot[projected] coordinates {(56.13, 6.13)};
                \label{plots:projected}
                \addplot[sphconv] coordinates {(8075.14, 63.54)};
                \label{plots:sphconv}
                \addplot[ktn] coordinates {(70.12, 69.48)};
                \label{plots:ktn}
                \coordinate (top) at (rel axis cs:0,1);

            \nextgroupplot[
                ymin=0,
                ymax=80,
                xmax=1000,
                xmode=log,
                only marks,
                ylabel={\large Accuracy},
                xlabel={\large Time (s) / Image},
            ]
                \addplot[equi] coordinates      {(0.02, 41.63)};
                \addplot[cubemap] coordinates   {(23.56, 49.29)};
                \addplot[s2cnn] coordinates     {(0.46, 4.32)};
                \addplot[sphcnn] coordinates    {(0.20, 6.06)};
                \addplot[spherenet] coordinates {(1.47, 46.68)};
                \addplot[sphunet] coordinates   {(2.39, 24.98)};
                \addplot[sphconv] coordinates   {(108.56, 63.54)};
                \addplot[projected] coordinates {(1.47, 6.13)};
                \addplot[ktn] coordinates       {(4.14, 69.48)};
                \coordinate (bot) at (rel axis cs:1,0);
        \end{groupplot}

        \path (top-|current bounding box.east) -- coordinate(legendpos) (bot-|current bounding box.east);
        \matrix[
            matrix of nodes,
            anchor=west,
            draw,
            inner sep=0.2em,
            ampersand replacement=\&,
            style={font=\large},
            column 2/.style={anchor=base west, font=\large},
        ]
        at ([xshift=0em, yshift=0em]legendpos) {
            \ref{plots:equi}        \& \textsc{Equirectangular} \\
            \ref{plots:cubemap}     \& \textsc{Cubemap} \\
            \ref{plots:s2cnn}       \& \textsc{$S^{2}CNN$}~\cite{cohen2017convolutional}\\
            \ref{plots:sphcnn}      \& \textsc{Spherical CNN}~\cite{sphericalcnn}\\
            \ref{plots:spherenet}   \& \textsc{SphereNet}~\cite{spherenet}\\
            \ref{plots:sphunet}     \& \textsc{Spherical U-Net}~\cite{saliency360video}\\
            \ref{plots:sphconv}     \& \textsc{SphConv}~\cite{su2017nips}\\
            \ref{plots:projected}   \& \textsc{Projected}\\
            \ref{plots:ktn}         \& \textsc{KTN} (Ours)\\
        };
    \end{tikzpicture}
}
    \vspace{-15pt}
    \caption{
        Model size (top) and speed (bottom) vs.~accuracy for VGG.
        \textsc{KTN} is orders of magnitude smaller than \textsc{SphConv},
        and it is similarly or more compact as all other models, while being significantly more accurate.
    }
    \label{fig:computational_cost}
    \vspace{-8pt}
\end{figure}
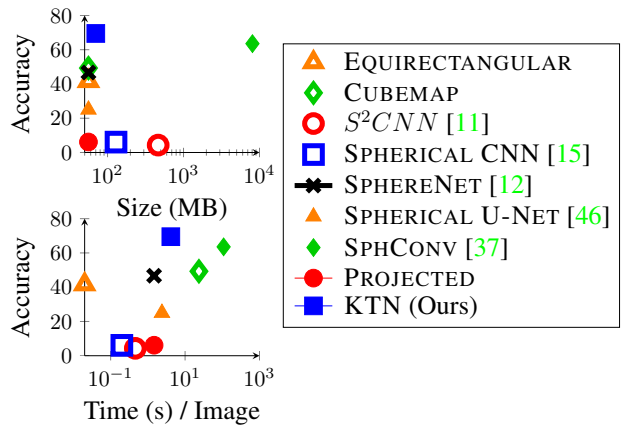

Fig.~\ref{fig:computational_cost} shows the results.
We can see that the model size of \textsc{KTN} is very similar to \textsc{Equirectangular}, \textsc{Cubemap} and \textsc{Projected}.
In fact, it is only $25\%$ (14MB) larger than the source CNN.
At the same time,
\textsc{KTN} achieves a much better accuracy compared with all the models that have a comparable size.
Compared with \textsc{SphConv},
\textsc{KTN} not only achieves a higher accuracy but is also orders of magnitude smaller.
Similarly, \textsc{$S^{2}$CNN} and \textsc{Spherical CNN} increase model size by $131\%$ and $727\%$ while performing worse in terms of accuracy.
Note that we do not include parameters that can be computed analytically,
such as the bases for \textsc{$S^{2}CNN$} and the projection matrices for \textsc{SphereNet},
though in practice they also add further memory overhead for those baselines.

On the other hand, the computational cost of \textsc{KTN} is naturally much higher than \textsc{Equirectangular}.
The latter only needs to run the source CNN on an equirectangular image,
whereas the convolution kernels are generated at run time for \textsc{KTN}.
However, as all the results show, \textsc{KTN} is much more accurate.
Furthermore, \textsc{KTN} is 26 times faster than \textsc{SphConv},
since the smaller model size allows the model to be evaluated on GPU.

\section{Conclusion}

We propose the Kernel Transformer Network for transfering CNNs from perspective images to $360\degree$ images.
KTN learns a function that transforms a kernel to account for the distortion in the equirectangular projection of $360\degree$ images.
The same KTN model can transfer to multiple source CNNs with the same architecture, 
significantly streamlining the process of visual recognition for $360\degree$ images.
Our results show KTN outperforms existing methods while providing superior scalability and transferability.

\vspace{0.1in}
\noindent {\bf Acknowledgement}.
We thank Carlos Esteves for the help on \textsc{Spherical CNN} experiments.
This research is supported in part by NSF IIS-1514118, an AWS gift, a Google PhD Fellowship, and a Google Faculty Research Award.

\appendix

The supplementary materials consist of:
\begin{enumerate}[leftmargin=*,label=\Alph*]
    \setlength{\itemsep}{1pt}
    \setlength{\parskip}{1pt}
    \item Complete architecture of \textsc{KTN}
    \item Experimental details
    \item Model transferability experiment on \emph{Pascal VOC}
    \item Comparison of model accuracy versus depth
    \item Discussion of multiple projections baselines
    \item Additional qualitative detection examples
\end{enumerate}

\section{\textsc{KTN} Architecture}

\begin{figure}[t]
    \center
    \includegraphics[width=\linewidth]{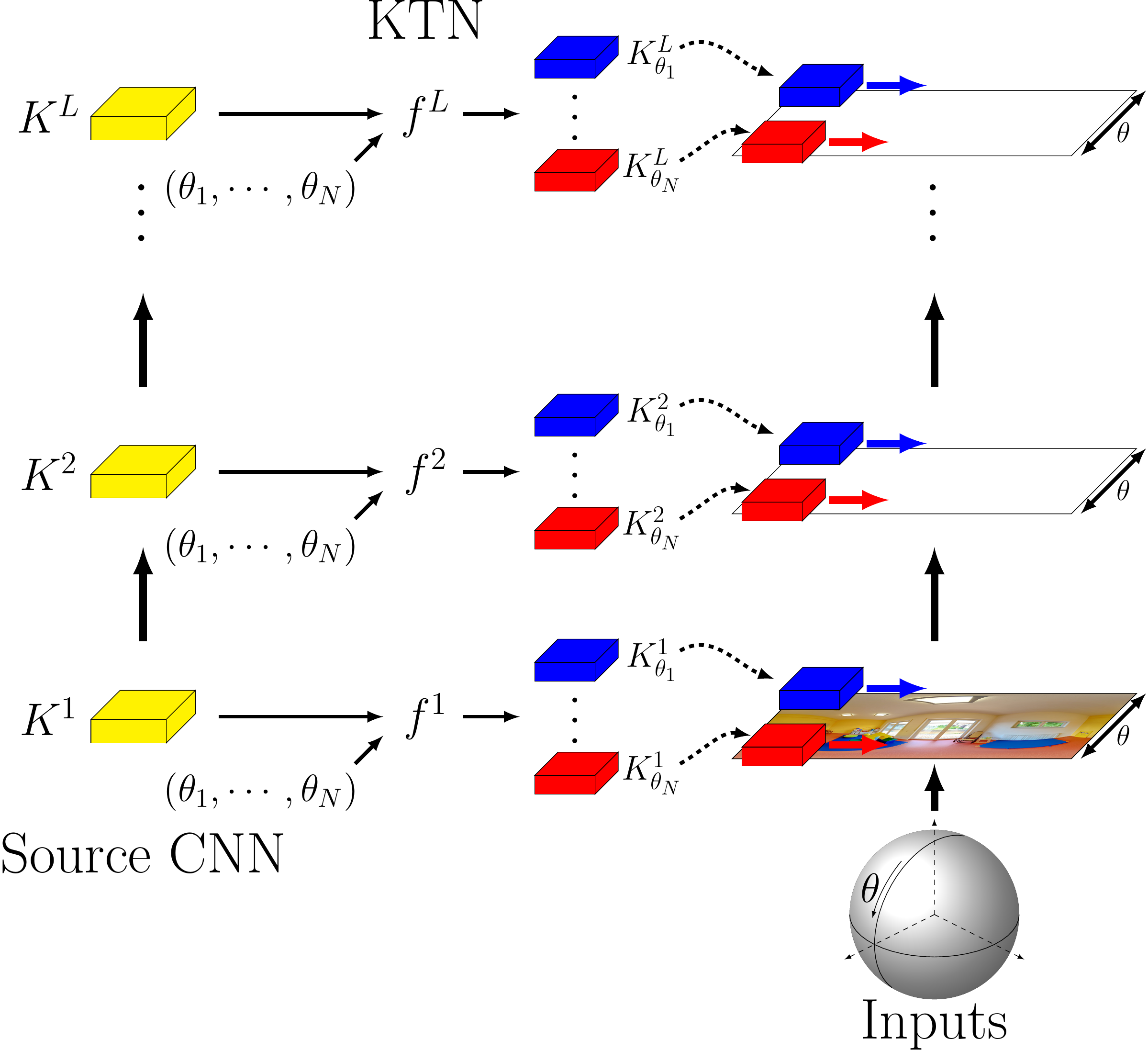}
    \vspace{-21pt}
    \caption{
        Application of spherical convolution using \textsc{KTN}.
    }
    \label{fig:ktn_net}
\end{figure}

\begin{figure}[t]
    \center
    \includegraphics[width=\linewidth]{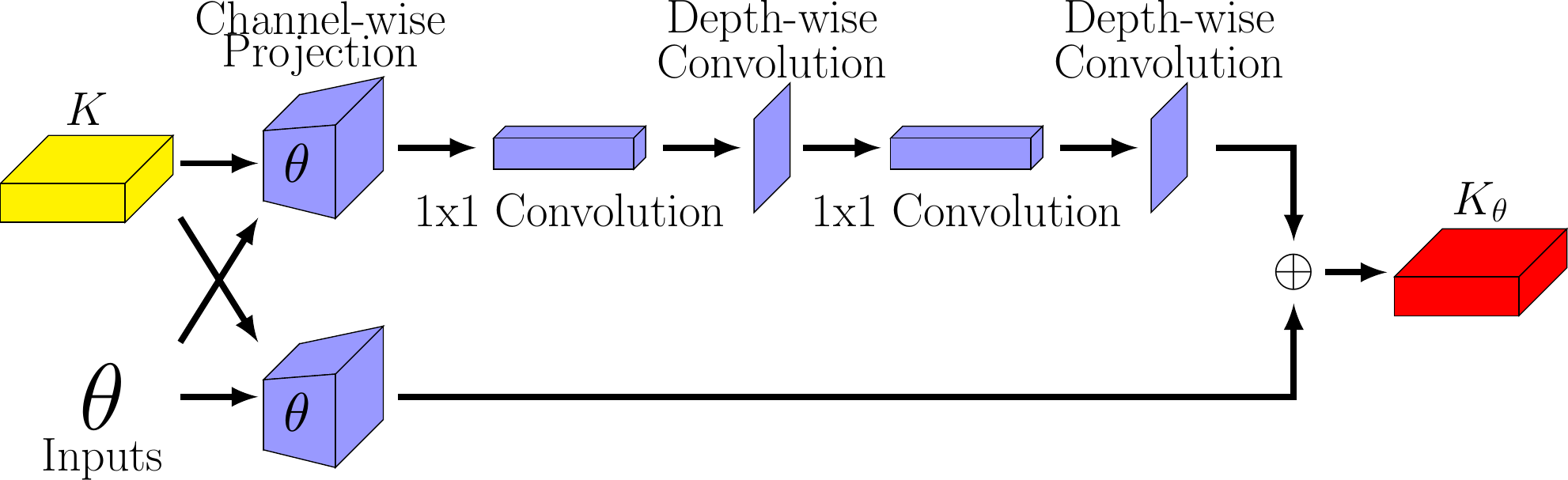}
    \vspace{-18pt}
    \caption{
        Full architecture of \textsc{KTN}.
    }
    \label{fig:ktn_arch}
    \vspace{-8pt}
\end{figure}

In this section, we show the complete architecture of \textsc{KTN}.
Fig.~\ref{fig:ktn_net} shows how to apply \textsc{KTN} for spherical convolution.
For each layer $l{\in}\{1,\cdots,L\}$ of the source CNN, we learn a function $f^{l}$ that transforms the source kernel $K^{l}$ to $K^{l}_{\theta_{i}}$ for every $\theta \in [0, \pi]$.
The output kernel $K^{l}_{\theta_{i}}$ is then applied to the $360\degree$ equirectangular image at the corresponding row of $\theta_{i}$.
We find that it is unnecessary to generate one kernel for each row in the equirectangular projection,
because spherical convolution kernels for adjacent rows are usually similar.
In practice, we share the same kernel every five rows to reduce the computational cost and model size.
Fig.~\ref{fig:ktn_arch} shows the full architecture of \textsc{KTN}.
\textsc{KTN} uses a ResNet-like architecture.
For both branches,
it uses a row dependent channel-wise projection to resize the kernel to the target size.
The residual branch then applies two depth separable convolution blocks before adding the output with that of the shortcut branch.
Each depth separable convolution block consists of ReLU-pointwise conv-ReLU-depthwise conv.

To compute the target kernel size at a given polar angle,
we first back project the receptive field of the source kernel to equirectangular projection.
The minimum bounding box centered at the polar angle that can cover the receptive field on equirectangular projection is then selected as the target kernel shape.
Note that we restrict the kernel height and width to be an odd number to ensure that the kernel is defined on the equirectangular pixel space.
Because the size of the back projected receptive field may grow rapidly and span the entire image,
we restrict the actual kernel width and height to be less then 65 pixels and dilate the kernel to increase the effective receptive field if necessary.

For Spherical Faster R-CNN,
we define the bounding boxes on the tangent plane: we project the features to the tangent plane and apply the RPN and detector networks there.
The implementation is the same as \textsc{SphConv}~\cite{su2017nips}.

\section{Experimental Details}
\vspace{-0.02in}

In this section, we describe additional experimental details that could not fit in the main paper.

\subsection{Datasets}
\label{sub:datasets}

The following is an expanded version of the dataset descriptions in the main text.

\emph{Spherical MNIST} is constructed from the MNIST dataset by backprojecting the digits into equirectangular projection.
The resolution of the resultant $360\degree$ image is $160{\times}80$, and the digit covers a $65.5\degree$ field-of-view (FOV).
For the training set, we project each digit to a random polar angle $\theta{\in}[0, \pi]$.
For the test set, we project each digit to nine different polar angles $\theta{=}8\degree, 16\degree, \ldots, 72\degree$,
which results in a test set that is nine times larger.
Note that we do not rotate the digit itself because digits are oriented by definition (e.g.~6 versus 9).
All baselines are trained to predict the digit label on the \emph{Spherical MNIST} training set except \textsc{SphConv}, which does not require such labels.
Both \textsc{KTN} and \textsc{SphConv} are trained to re-produce the top-most convolution output (conv3).
Classification accuracy averaged across $\theta$ is used as the evaluation metric.

\emph{Pano2Vid} is a real world $360\degree$ video dataset~\cite{su2016accv}.
It contains 86 videos from four categories: ``Hiking,'' ``Parade,'' ``Soccer,'' and ``Mountain Climbing.''
Following~\cite{su2017nips},
we sample 1,056 frames from the first three categories for training and 168 frames from the last category for testing,
and the frames are resized to $640{\times}320$ resolution.
The root-mean-square error (RMSE) over the final convolution outputs is used as the evaluation metric.

\emph{Pascal VOC} is a perspective image dataset with object annotations.
Similar to \emph{Spherical MNIST},
we backproject the object bounding boxes to equirectangular projection but with $640{\times}320$ resolution.
Each bounding box is projected to different polar angles $\theta{\in}\{18\degree, 36\degree, 54\degree, 72\degree, 90\degree\}$ and covers a $65.5\degree$ FOV.
Because the perspective images do not cover the full $360\degree$ FOV, regions outside the FOV of the original image are zero-padded (black).
This dataset is used for evaluation only.
Following the experiment setting of the Faster R-CNN~\cite{ren2015fasterRCNN} source model,
we evaluate all methods on the validation set of Pascal VOC 2007.
We use the accuracy of the detector network in Faster R-CNN as the evaluation metric.
The ground truth bounding box is used for ROI-pooling during evaluation for all methods.

\subsection{Baselines}
\label{sub:baselines}

In this section,
we expand on the implementation details of each baseline method.
We keep the number of layers and kernels the same for all methods.
For the \emph{Spherical MNIST} dataset,
the models consist of three convolution layers followed by a max-pooling over the spatial dimensions and a fully connected layer.
The convolution layers have 32, 64, and 128 kernels respectively,
and the resolution of the feature map is reduced by a factor of two using max-pooling after each convolution layer.
For the \emph{Pano2Vid} and \emph{Pascal VOC} datasets,
the models have the same number of layers and kernels as the VGG16 architecture.
Following \textsc{SphConv}~\cite{su2017nips},
we remove the max-pooling operation in the network and use dilated convolution with factor of two in the conv5 layers to increase the receptive field.
The differences between different methods are in the convolution and pooling operations as described below.

\begin{itemize}[leftmargin=*,label=$\bullet$]
    \item \textsc{Equirectangular}---Apply ordinary CNNs on the $360\degree$ image in its equirectangular projection.
    \item \textsc{Cubemap}---Apply ordinary CNNs on the $360\degree$ image in its cubemap projection,
        with cube padding~\cite{cheng2018cubepadding}.
        For the \textsc{Pano2Vid} and \textsc{Pascal VOC} datasets,
        the conv5\_3 feature map is re-projected to equirectangular projection as the final output.
    \item \textsc{$S^{2}CNN$}~\cite{cohen2017convolutional}---We use the S2Convolution and SO3Convolution in the authors' implementation\footnote{\url{https://github.com/jonas-koehler/s2cnn}} for convolution.
        S2Convolution is applied in the first convolution layer, and SO3Convolution is used for the other layers.
        The default near identity grid is used for both S2 and SO3 convolution.
        Furthermore, we reduce the feature map resolution by reducing the output bandwidth instead of using max-pooling following the authors' implementation.
        The input resolution is $80{\times}80$ for \emph{Spherical MNIST} and $64{\times}64$ for \emph{Pano2Vid} and \emph{Pascal VOC}.
        For \emph{Spherical MNIST}, we use SO(3) integration instead of max-pooling to reduce the final feature map.
        For \emph{Pano2Vid} and \emph{Pascal VOC}, because the output of SO3Convolution is a 3D feature map,
        we add a 1x1 convolution layer on top of the conv5\_3 output to generate a 2D feature map.
        The feature map is then resized to $640{\times}320$ as the final output.
        We reduce the output bandwidth in conv2\_2 and conv3\_3 and distribute the model to four NVIDIA V100 GPUs using model parallelism due to the GPU memory limit.
    \item \textsc{Spherical CNN}~\cite{sphericalcnn}---We use the sphconv module in the authors' implementation\footnote{\url{https://github.com/daniilidis-group/spherical-cnn}} for convolution.
        Similar to \textsc{$S^{2}CNN$}, we replace max-pooling with spectral pooling.
        Furthermore, we apply batch normalization in each convolution layer following the example code.
        The input resolution is $80{\times}80$ for all datasets.
        For the \emph{Pano2Vid} and \emph{Pascal VOC} dataset, we reduce the output bandwidth in conv4\_1 and conv5\_1 due to the memory limit.
        The conv5\_3 feature map is resized to $640{\times}320$ as the final output.
    \item \textsc{Spherical U-Net}~\cite{saliency360video}---We use the SphericalConv module in Spherical U-Net\footnote{\url{https://github.com/xuyanyu-shh/Saliency-detection-in-360-video}} for convolution.
        We apply batch normalization and set the kernel size to $8{\times}4$ following the authors' example.
        For the \emph{Pano2Vid} and \emph{Pascal VOC} dataset, the input is resized to $160{\times}80$ due to memory limit,
        and the conv5\_3 feature map is resized to $640{\times}320$ as the final output.
        The model is distributed to four NVIDIA V100 GPUs using model parallelism.
    \item \textsc{SphereNet}~\cite{spherenet}---We implement the \textsc{SphereNet} model using row dependent channel-wise projection.  The authors' code and data were unavailable at the time of submission.
        Because feature projection is the weighted sum of the features,
        the projection weights can be combined with the kernel weights as a single kernel.
        We derive the weights of the channel-wise projection using the feature projection operation and train the source kernels.
        For the \emph{Pano2Vid} dataset, 
        we train each layer independently using the same objective function as \textsc{KTN} because the entire model cannot fit in GPU memory.
    \item \textsc{SphConv}~\cite{su2017nips}---We use the authors' implementation\footnote{\url{https://github.com/sammy-su/Spherical-Convolution}}.
        Because the model is too large to fit into GPU memory even for evaluation, it is run on CPUs.\footnote{For the other baselines, testing is still possible with GPUs.}
    \item \textsc{Projected}---Assuming that the kernel transformation $f$ can be modeled using the tangent plane-to-sphere projection,
        we derive the analytic solution for the kernels $K_{\theta}$ using bilinear interpolation.        
\end{itemize}
Note that the aspect ratio for the inputs is 1:1 for \textsc{$S^{2}CNN$} and \textsc{Spherical CNN}.
This is the requirement of the methods, so we reduce the resolution along the azimuthal angle.
The input aspect ratio for all other methods is 2:1 following the common format of $360\degree$ images.

\begin{figure}[t]
    \center
    \includegraphics[width=\linewidth]{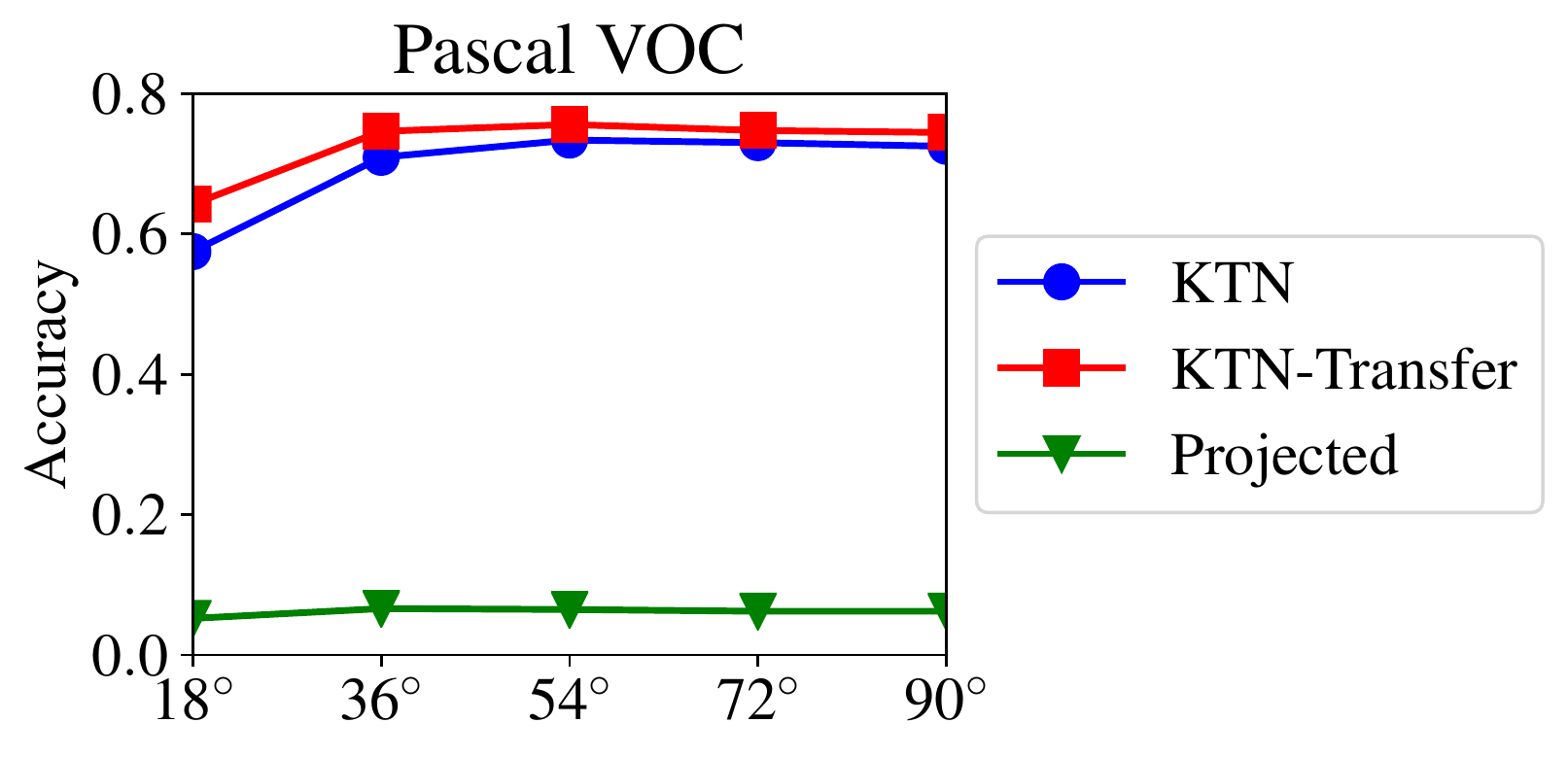}
    \vspace{-18pt}
    \caption{
        Model transferability on \emph{Pascal VOC}.
    }
    \label{fig:transferability}
\end{figure}

\begin{figure}[t]
    \center
    \includegraphics[width=\linewidth]{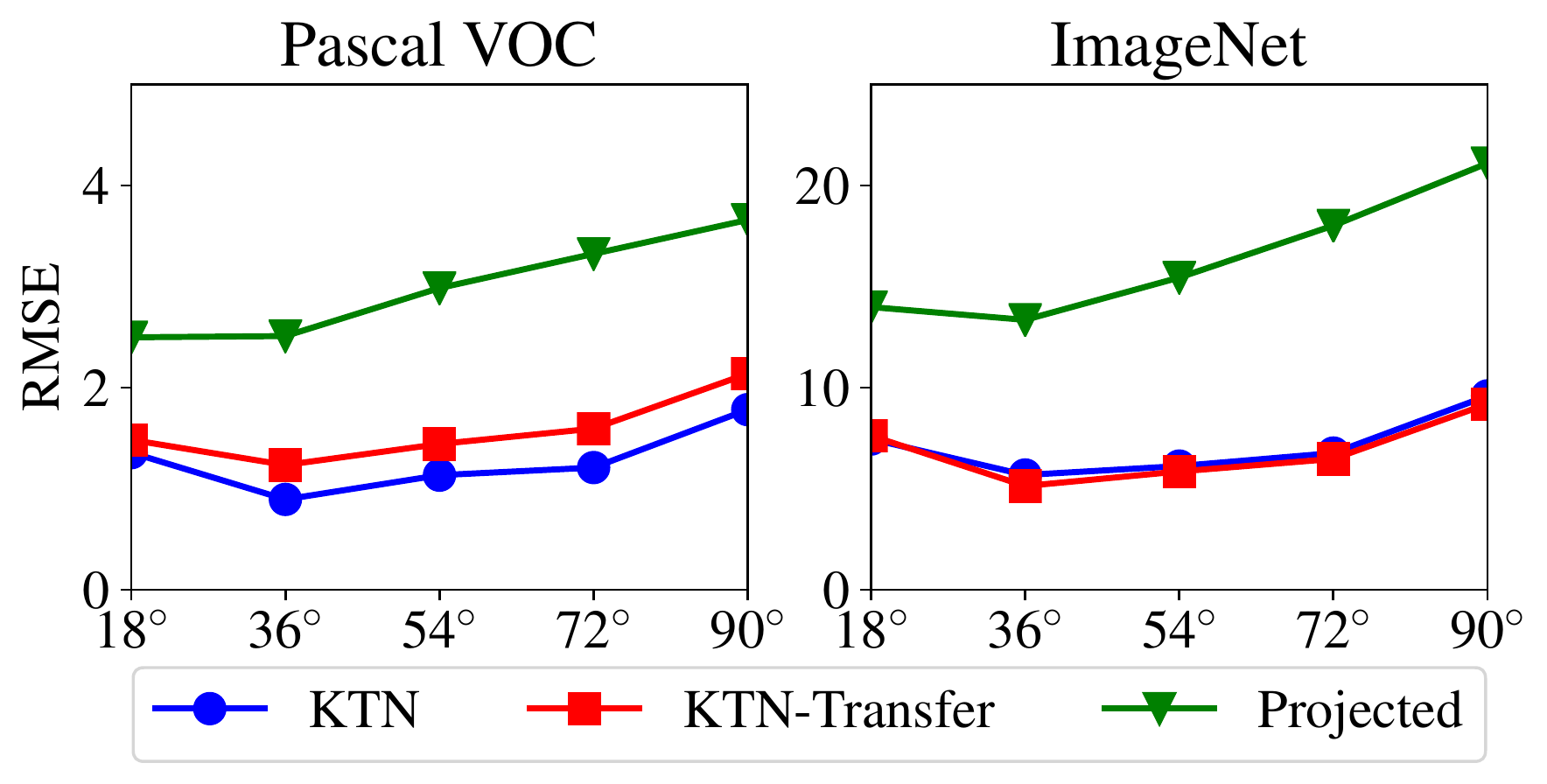}
    \vspace{-18pt}
    \caption{
        Transferability of ImageNet trained VGG.
    }
    \label{fig:imagenet_transferability}
    \vspace{-4pt}
\end{figure}

\subsection{Training Details}
\label{sub:implementation}

We train all the methods using ADAM~\cite{kingma2014adam} for 40 epochs. 
The learning rate is initialized to $1.0{\times}10^{-3}$ and is decreased by a factor of 10 after 20 epochs.
We also apply L2 regularization with weight $5{\times}10^{-4}$.
For \emph{Pano2Vid},
the batch size is set to one for \emph{$S^{2}CNN$}, two for \emph{Spherical CNN}, and four for all other methods,
which is again limited by the memory.
For \emph{Spherical MNIST}, the batch size is set to 64 for all methods except \textsc{$S^{2}CNN$}, which uses a batch size of 16.
The weights are randomly initialized using a normal distribution with standard deviation $0.01$.
The training time for \textsc{KTN} on \emph{Pano2Vid} is about a week using six AWS p3.8xlarge instances with V100 GPUs.

\section{Transferability on \emph{Pascal VOC}}

As noted in the main paper, in this section, we evaluate the transferability of \textsc{KTN} on \emph{Pascal VOC}.
In particular, we measure whether the \textsc{KTN} model trained on a VGG source model can be applied to Faster R-CNN to perform object detection.
The result is in Fig.~\ref{fig:transferability}.
Again, \textsc{KTN} performs almost identical regardless of the source model on which it is trained.
We also evaluate the transferability between VGG trained for ImageNet classification and Faster R-CNN trained for Pascal object detection.
The result is in Fig.~\ref{fig:imagenet_transferability}.
The results are consistent with that in Sec.~4.3 of the main paper and verifies that \textsc{KTN} is transferable across source CNNs with the same architecture.

\section{Model Accuracy versus Depth}

As discussed in the main paper, the interpolation assumption made by \textsc{SphereNet}~\cite{spherenet} and the \textsc{Projected} baselines is problematic, particularly at deeper layers as errors accumulate.
Hence, we compare the accuracy of \textsc{SphereNet}~\cite{spherenet}, \textsc{Projected}, and \textsc{KTN} with different network depths.
We change the network depth by feeding in the ground truth value of the intermediate layer and compare the RMSE of conv5\_3 outputs.
The experiment is performed on \emph{Pano2Vid} using Faster R-CNN source model.

\begin{figure}[t]
    \center
\resizebox{\linewidth}{!} {
    \begin{tikzpicture}
        \tikzset{mark options={mark size=4, solid}}
        \begin{axis}[
                width=0.75\linewidth,
                axis lines=left,
                ylabel={\large RMSE},
                xlabel={\large Depth},
                xlabel shift=-4pt,
            ]
            \pgfplotstableread[col sep=comma]{index.dat}\loadedtable
            \addplot[spherenet] table [x=Depth, y=spherenet] {\loadedtable};
            \label{plots:spherenet}
            \addplot[projected] table [x=Depth, y=projected] {\loadedtable};
            \label{plots:projected}
            \addplot[ktn] table [x=Depth, y=ktn] {\loadedtable};
            \label{plots:ktn}
            
        \end{axis}
        \coordinate (top) at (rel axis cs:0,1);
        \coordinate (bot) at (rel axis cs:1,0);
        \path (top-|current bounding box.east) -- coordinate(legendpos) (bot-|current bounding box.east);
        \matrix[
            matrix of nodes,
            anchor=west,
            draw,
            inner sep=0.2em,
            ampersand replacement=\&,
            style={font=\large},
            column 2/.style={anchor=base west, font=\large},
        ]
        at ([xshift=0em, yshift=0em]legendpos) {
            \ref{plots:spherenet}   \& \textsc{SphereNet}~\cite{spherenet}\\
            \ref{plots:projected}   \& \textsc{Projected}\\
            \ref{plots:ktn}         \& \textsc{KTN} (Ours)\\
        };
    \end{tikzpicture}
}
    \vspace{-18pt}
    \caption{
        Model accuracy at different layers.
    }
    \label{fig:layer_accuracy}
\end{figure}
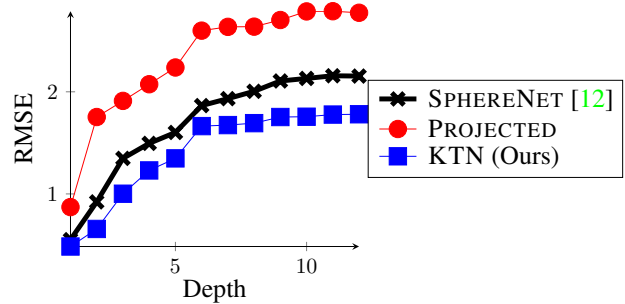

The results are in Fig.~\ref{fig:layer_accuracy}.
Not surprisingly, the error increases as the model depth increases for all methods.
More importantly, the gap between \textsc{KTN} and the other methods increases as the network becomes deeper.
The results suggest that the error of interpolated features increases as the number of non-linearities increases and is consistent with the analysis in Sec.~3.4 in the main paper.

\section{Multiple Projections Baseline}

\begin{figure}[t]
    \center
    \includegraphics[width=\linewidth]{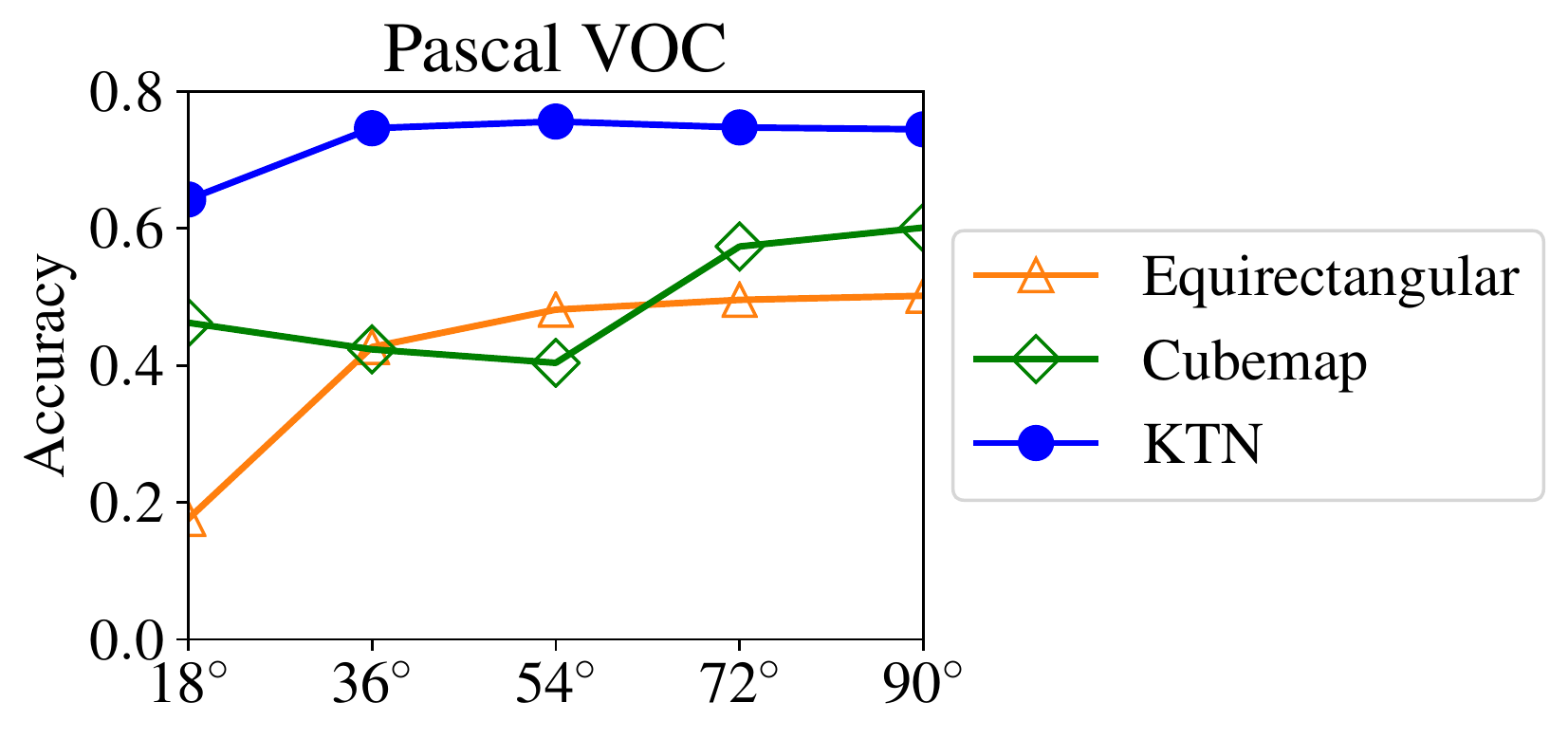}
    \vspace{-18pt}
    \caption{
        Model accuracy of projection based baselines.
    }
    \label{fig:multiprojection}
    \vspace{-4pt}
\end{figure}

Fig.~\ref{fig:multiprojection} shows that the worst accuracy of \textsc{KTN} (at $\theta{=}18\degree$) outperforms the best accuracy of \textsc{Equirectangular} and \textsc{Cubemap} (at $\theta{=}90\degree$).
While a possible method for improving the performance of the projection based methods
(i.e.~\textsc{Equirectangular} and \textsc{Cubemap}) is to aggregate the detection results from multiple projections to reduce the effect of distortion,
the results suggest that \textsc{Equirectangular} and \textsc{Cubemap} is less accurate then \textsc{KTN} even if they are always evaluated on the less distorted region.
This implies that \textsc{KTN} will always be more accurate than \textsc{Equirectangular} and \textsc{Cubemap} no matter how many different projections we sample.
Furthermore, evaluating the model on multiple projections increases the computational cost and introduces the problem of how to combine detection results,
which is non-trivial especially in dense prediction problems such as depth prediction.

\section{Object Detection Examples}

\begin{figure}[t]
    \center
    \includegraphics[width=\linewidth]{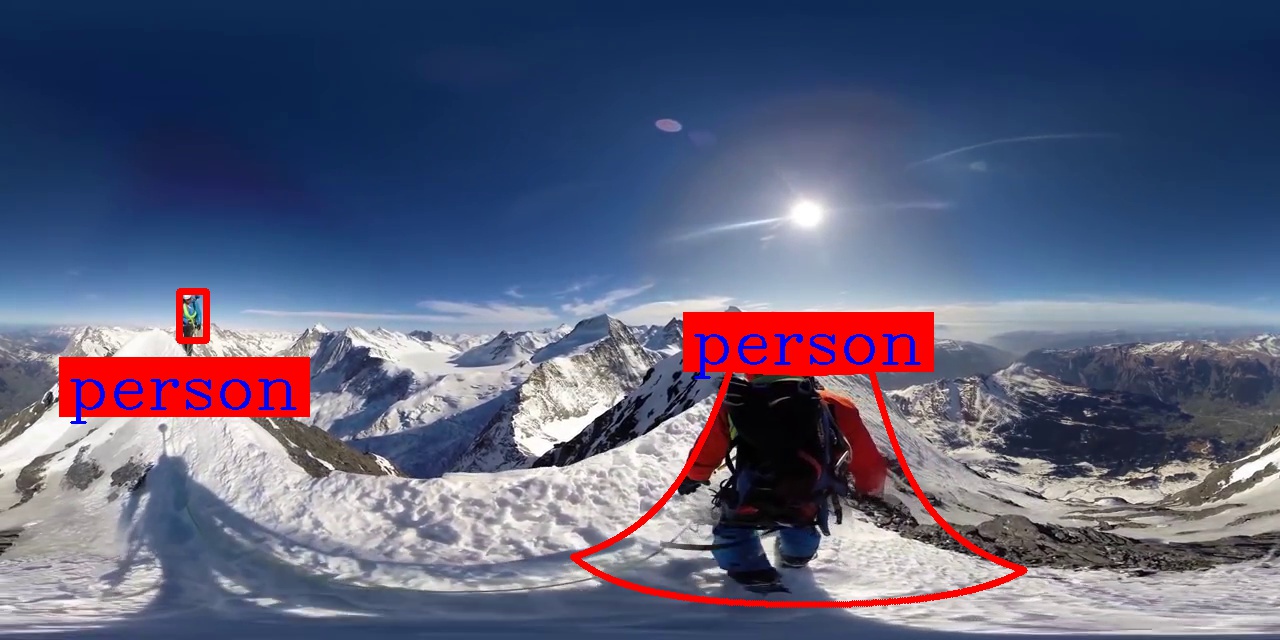}
    \vspace{1pt}
    \includegraphics[width=\linewidth]{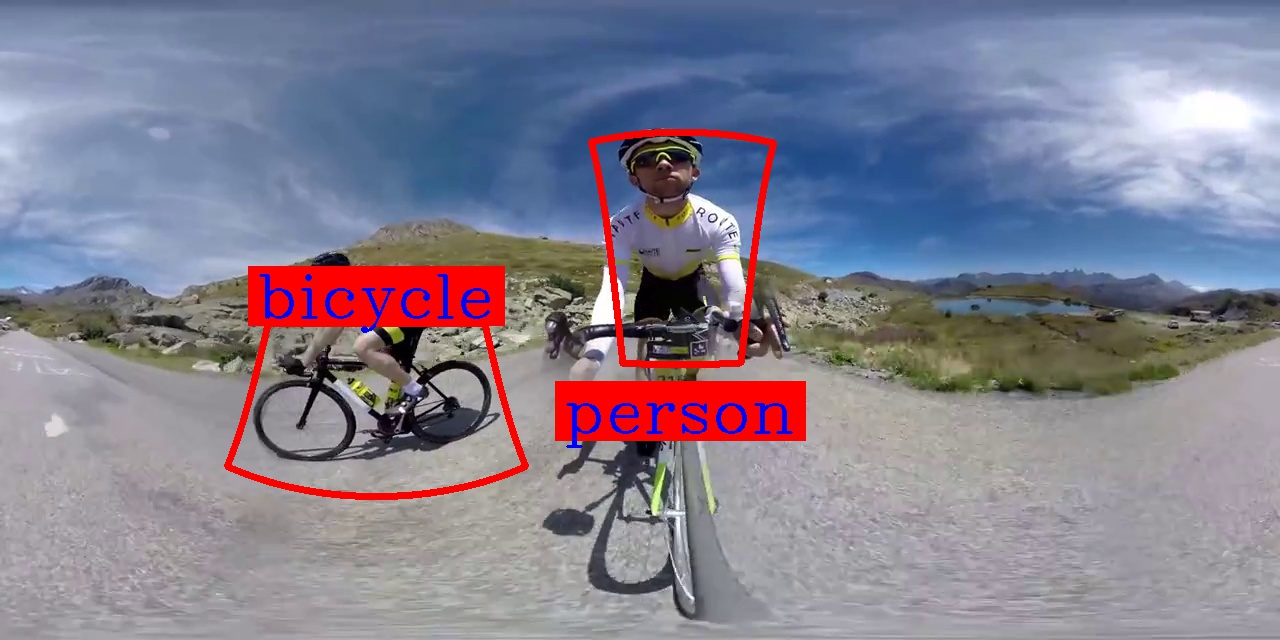}
    \vspace{1pt}
    \includegraphics[width=\linewidth]{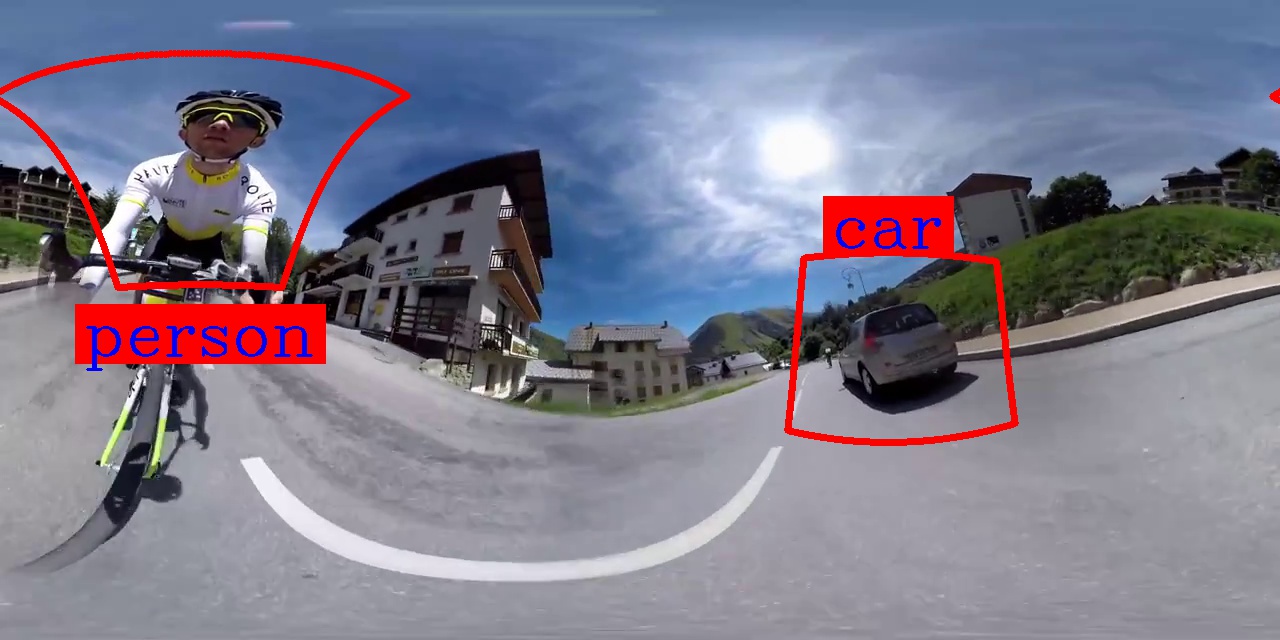}
    \vspace{1pt}
    \includegraphics[width=\linewidth]{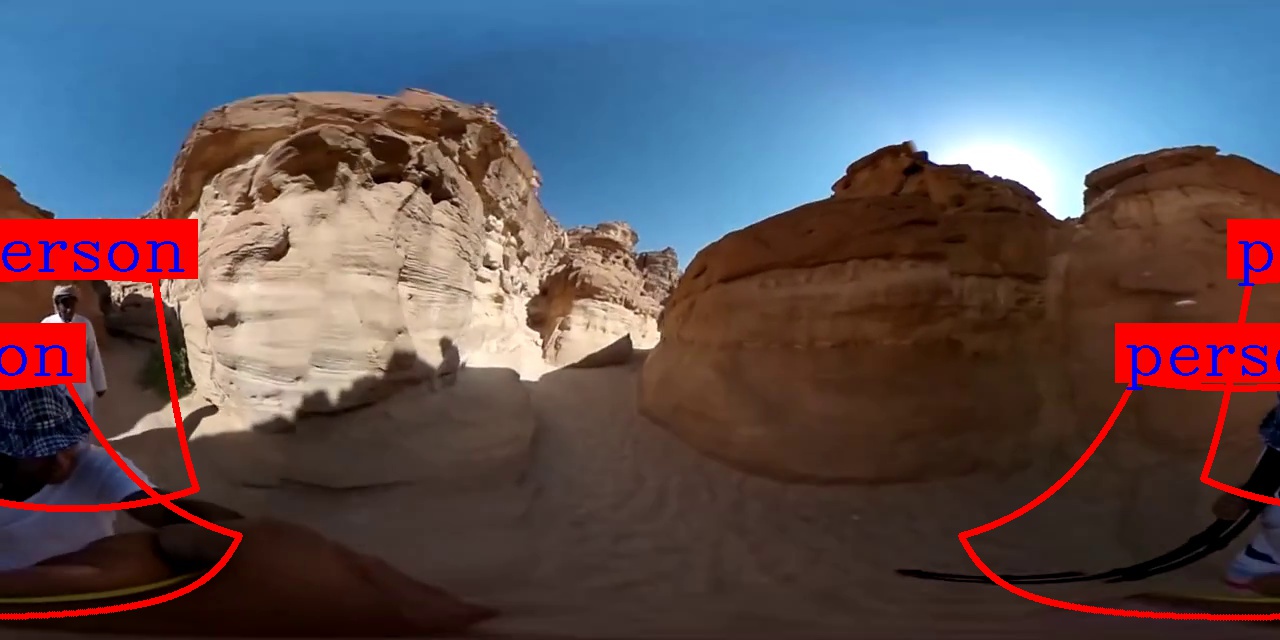}
    \vspace{1pt}
    \includegraphics[width=\linewidth]{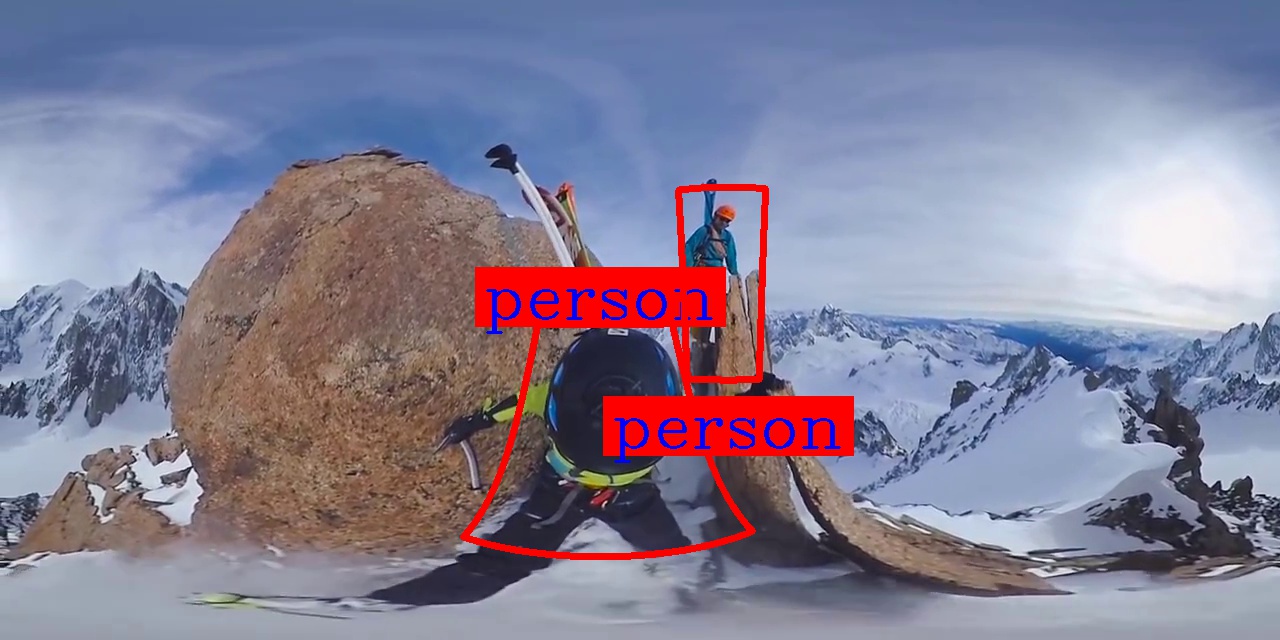}
    \vspace{-15pt}
    \caption{
        Object detection examples on \emph{Pano2Vid}.
    }
    \label{fig:pano2vid}
    \vspace{12pt}
\end{figure}

\begin{figure}[t]
    \center
    \includegraphics[width=\linewidth]{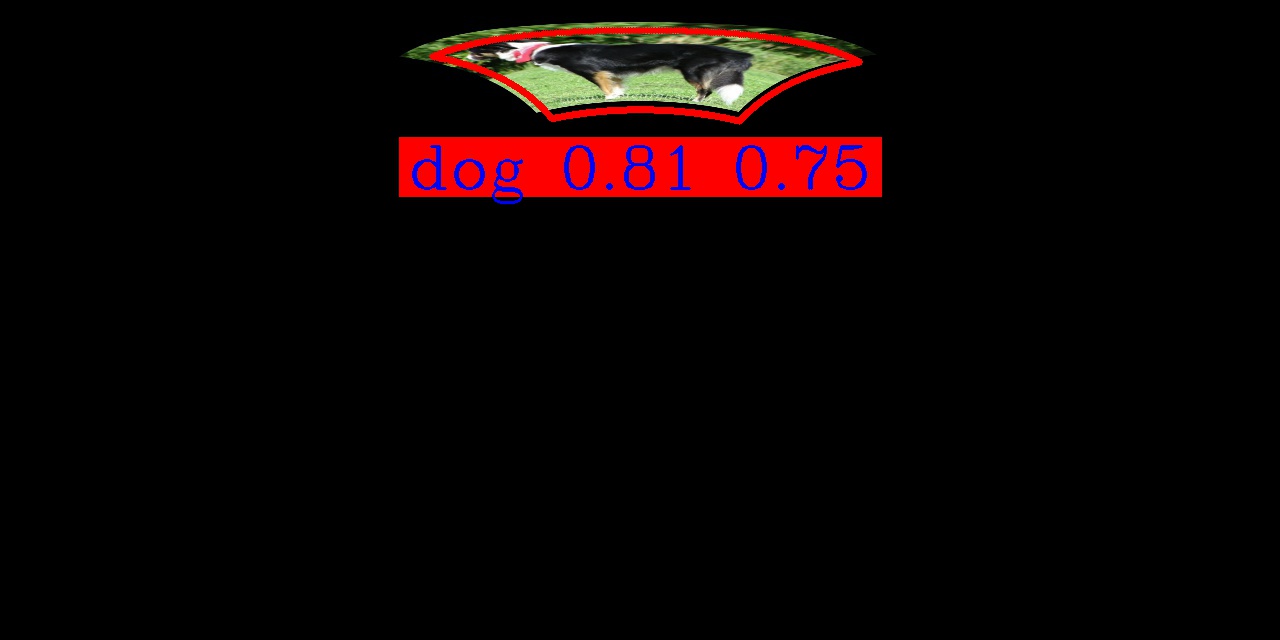}
    \vspace{1pt}
    \includegraphics[width=\linewidth]{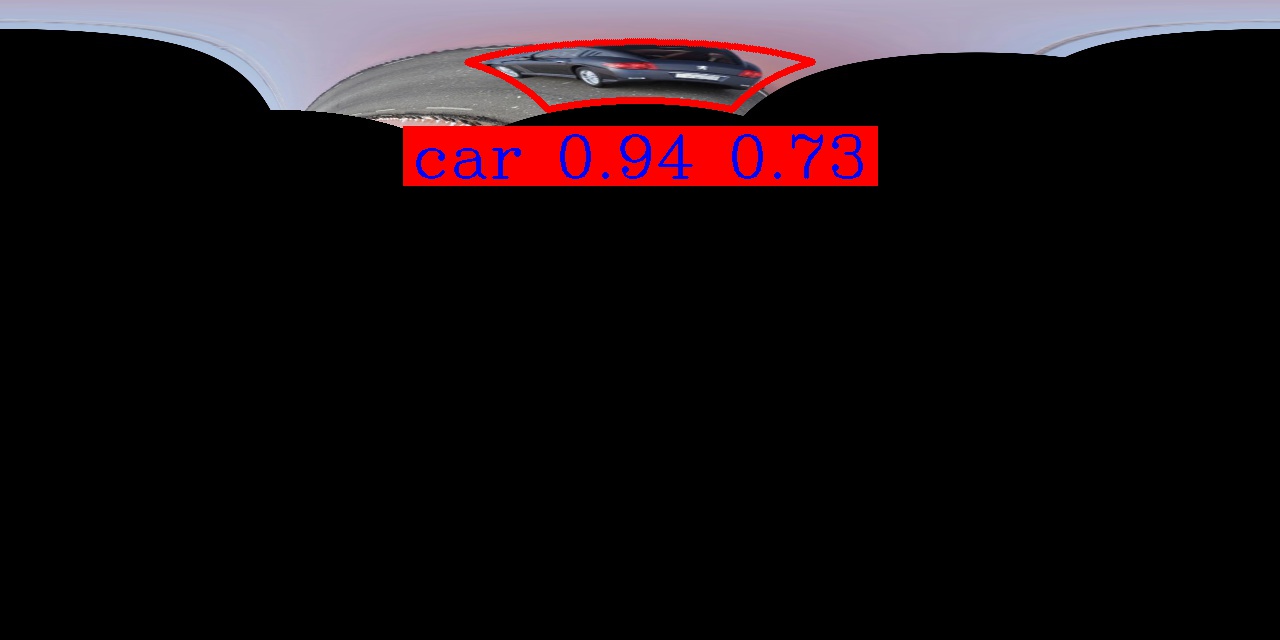}
    \vspace{1pt}
    \includegraphics[width=\linewidth]{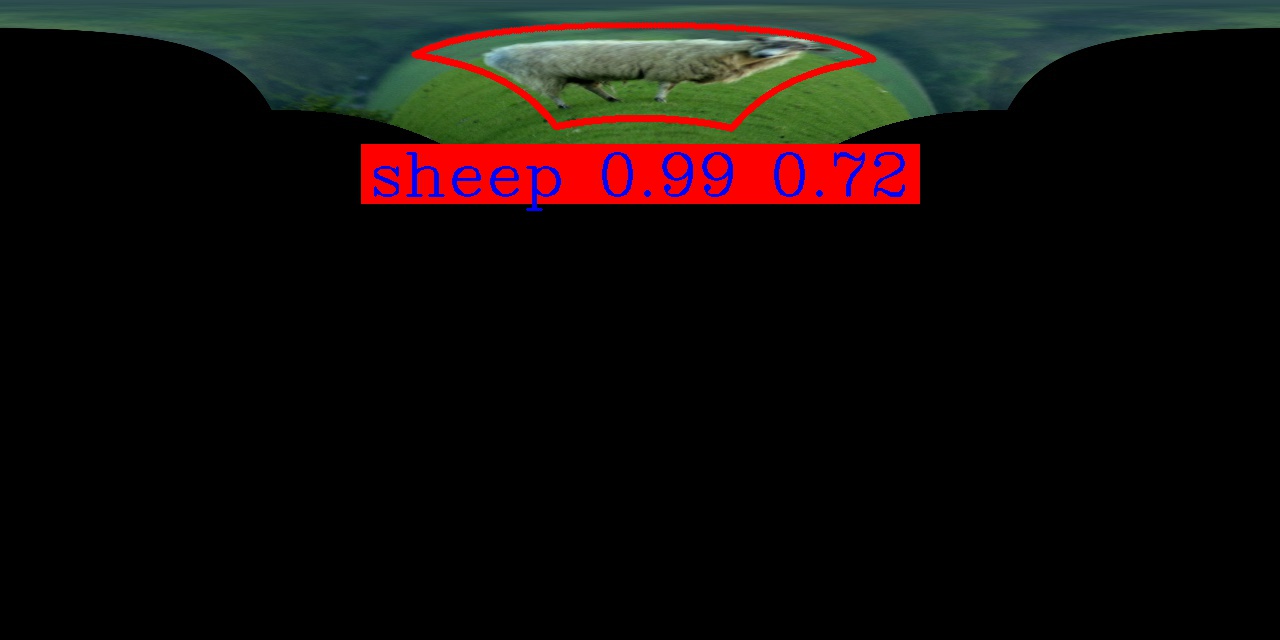}
    \vspace{1pt}
    \includegraphics[width=\linewidth]{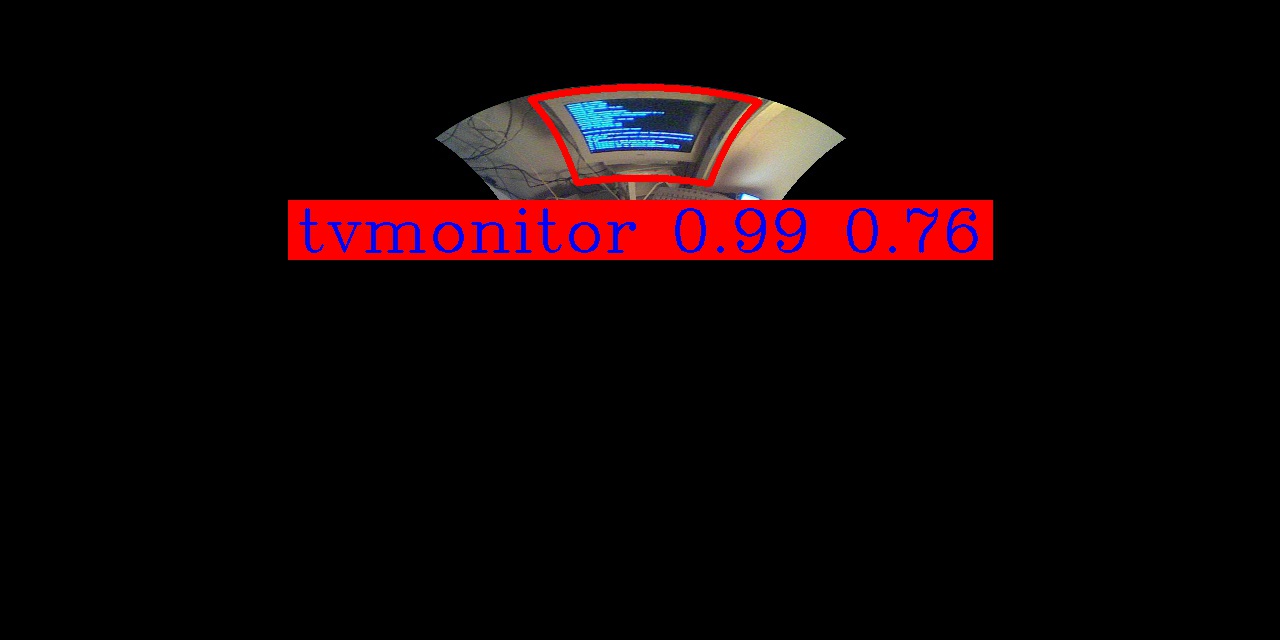}
    \vspace{1pt}
    \includegraphics[width=\linewidth]{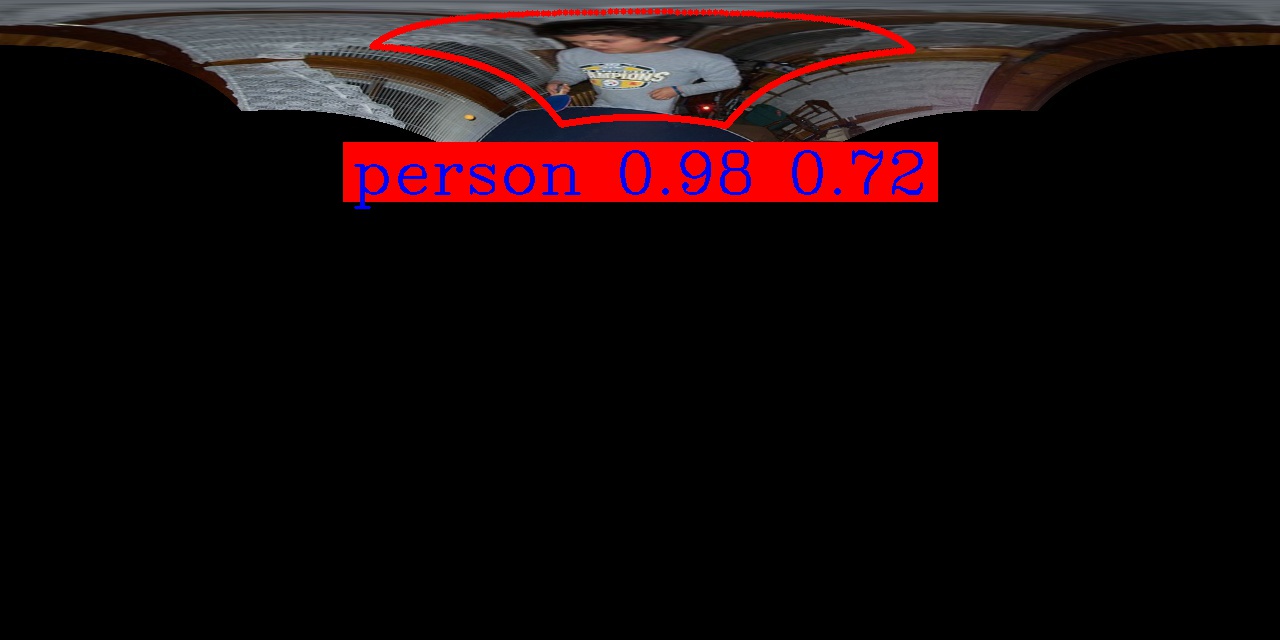}
    \vspace{-15pt}
    \caption{
        Object detection examples on 360-ified \emph{Pascal VOC images}.
    }
    \label{fig:pascal}
\end{figure}

In this section, we show additional object detection examples.
Fig.~\ref{fig:pano2vid} and Fig.~\ref{fig:pascal} show object detection examples on the \emph{Pano2Vd} and \emph{Pascal VOC} dataset, respectively.
Notice how KTN can detect the distorted objects by translating the source CNN appropriately to the spherical data.

\begin{figure}[t]
    \center
    \includegraphics[width=\linewidth]{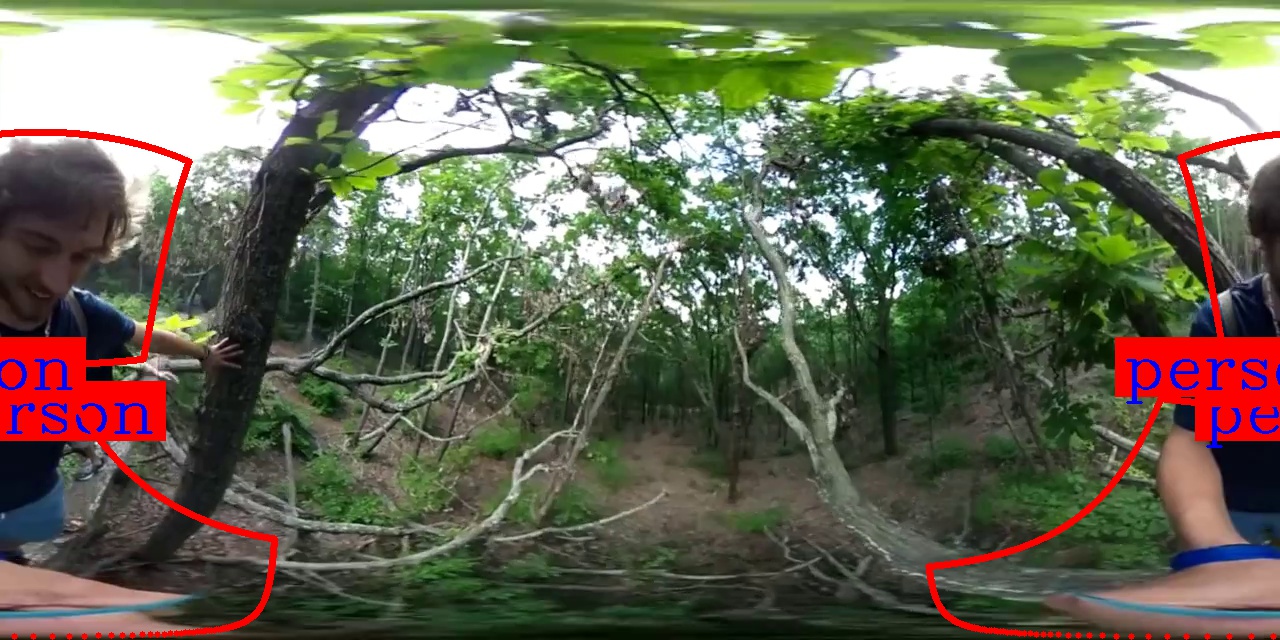}
    \includegraphics[width=\linewidth]{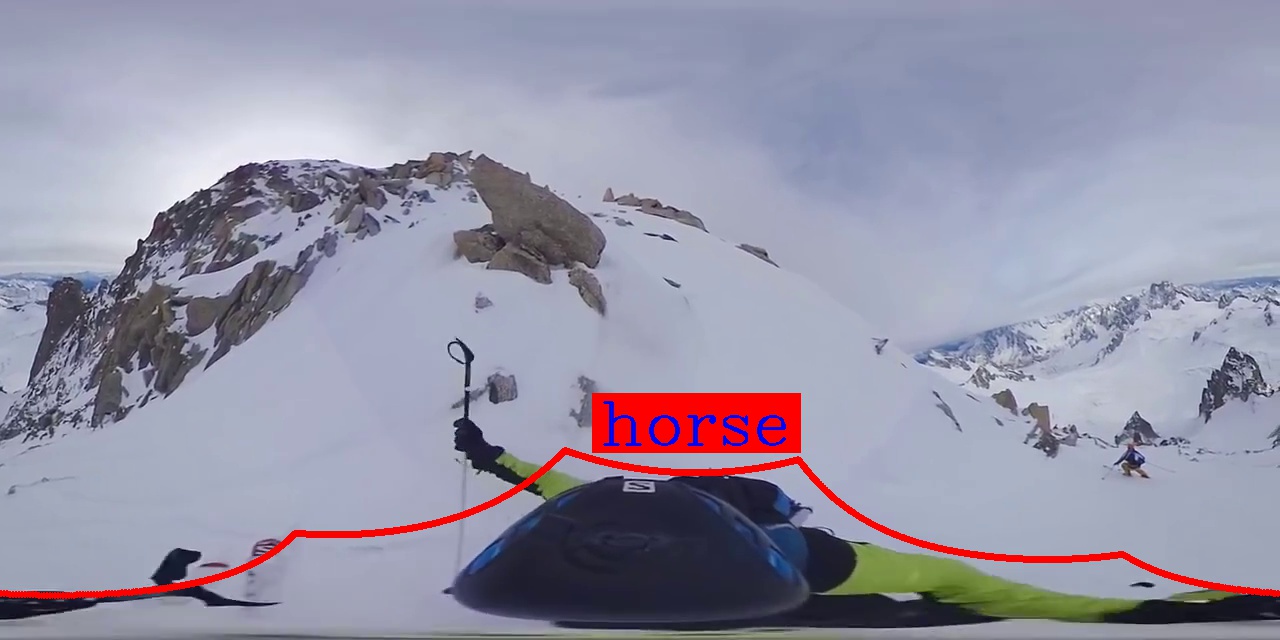}
    \vspace{-18pt}
    \caption{
        Failure cases on \emph{Pano2Vid}.
    }
    \label{fig:failures}
    \vspace{-12pt}
\end{figure}

Fig.~\ref{fig:failures} show failure examples on the \emph{Pano2Vid} dataset.
In the first example, the model fails to capture the entire human body and returns two positive detections instead of one.
This is caused by the fact that our method cannot handle close objects that cannot be captured by the FOV of perspective images.
In the second example, the model fails on the top view of the person because a top view is very rare in ordinary images.
The result indicates that the data distribution is different in $360\degree$ images and perspective images.  The performance of the model may be further improved if we can train the source CNNs on $360\degree$ images.

{\small
\bibliographystyle{ieee}
\bibliography{ktn}
}

\end{document}